\documentclass{article}

\usepackage{fullpage}
\usepackage[T1]{fontenc}
\usepackage{hyperref}
\usepackage{graphicx}
\usepackage{float}
\usepackage[backend=biber, style=alphabetic, citestyle=authoryear]{biblatex}
\usepackage[table, dvipsnames]{xcolor}
\usepackage{array}
\usepackage{caption}
\usepackage{multirow}
\usepackage{amsmath}
\usepackage{amssymb}
\usepackage{multicol}

\usepackage{tablefootnote}

\usepackage[bottom]{footmisc}

\title{
    \includegraphics[width=0.4\textwidth]{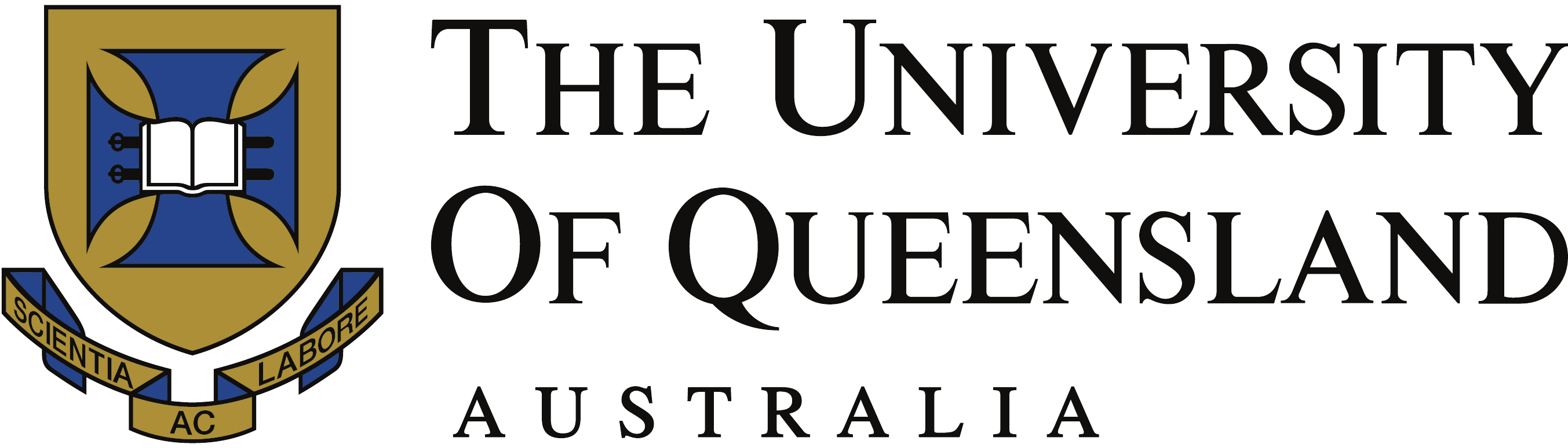}\\
    \vfill
    Cyber Attack Detection\\
    thanks to Machine Learning Algorithms\\[3mm]
    \large COMS7507: Advanced Security\\
    \vfill
}

\author{
    Antoine Delplace\\
    \href{mailto:a.delplace@uq.net.au}{a.delplace@uq.net.au}\\
    University of Queensland
    \and
    Sheryl Hermoso\\
    \href{mailto:s.hermoso@uq.net.au}{s.hermoso@uq.net.au}\\
    University of Queensland
    \and
    Kristofer Anandita\\
    \href{mailto:k.anandita@uq.net.au}{k.anandita@uq.net.au}\\
    University of Queensland
}

\date{May 17, 2019}

\begin{document}
\maketitle
\vfill

\begin{abstract}
    Cybersecurity attacks are growing both in frequency and sophistication over the years. This increasing sophistication and complexity call for more advancement and continuous innovation in defensive strategies. Traditional methods of intrusion detection and deep packet inspection, while still largely used and recommended, are no longer sufficient to meet the demands of growing security threats. \\
    
    As computing power increases and cost drops,  Machine Learning is seen as an alternative method or an additional mechanism to defend against malwares, botnets, and other attacks. This paper explores Machine Learning as a viable solution by examining its capabilities to classify malicious traffic in a network.\\
    
    First, a strong data analysis is performed resulting in 22 extracted features from the initial Netflow datasets. All these features are then compared with one another through a feature selection process.\\
    
    Then, our approach analyzes five different machine learning algorithms against NetFlow dataset containing common botnets. The Random Forest Classifier succeeds in detecting more than 95\% of the botnets in 8 out of 13 scenarios and more than 55\% in the most difficult datasets.\\
    
    Finally, insight is given to improve and generalize the results, especially through a bootstrapping technique.\\

    \textit{\underline{Useful keywords:} Botnet, Malware Detection, Cyber Attack Detection, NetFlow, Machine Learning}\\
    
    \textit{\underline{GitHub repository:}} \texttt{\href{https://github.com/antoinedelplace/Cyberattack-Detection}{https://github.com/antoinedelplace/Cyberattack-Detection}}
    
\end{abstract}

\vfill
\begin{center}
    \large
    \underline{Lecturer:}
    \href{mailto:m.portmann@uq.edu.au}{Marius Portmann}
\end{center}
\newpage

\tableofcontents
\newpage

\section{Introduction and Context}

Cybersecurity is evolving and the rate of cybercrime is constantly increasing. Sophisticated attacks are considered as the new normal as they are becoming more frequent and widespread. This constant evolution also calls for innovation in the cybersecurity defense. \\

There are existing solutions and a combination of these methods are still widely used. Network Intrusion Detection and Prevention Systems (IDS/IPS) monitor for malicious activity or policy violations. Signature-based IDS relies on known signatures and is effective at detecting malwares that match these signatures. Behaviour-based IDS, on the other hand, learns what is normal for a system and reports on any trigger that deviates from it. Both types, though effective, have some weaknesses. Signature-based systems rely on signatures of known threats and thus ineffective for zero-day attacks or new malware samples. Traditional behaviour-based systems rely on a standard profile which is hard to define with the growing complexity of networks and applications, and thus may be ineffective for anomaly detection. Full data packet analysis is another option, however, it is both computationally expensive and risks exposure of sensitive user information. \\

Machine Learning (ML) has gained a wide interest on many applications and fields of study, particularly in Cybersecurity. With hardware and computing power becoming more accessible, machine learning methods can be used to analyze and classify bad actors from a huge set of available data. There are hundreds of Machine Learning algorithms and approaches, broadly categorized into supervised and unsupervised learning. Supervised learning approaches are done in the context of Classification where input matches to an output, or Regression where input is mapped to a continuous output. Unsupervised learning is mostly accomplished through Clustering and has been applied to exploratory analysis and dimension reduction. Both of these approaches can be applied in Cybersecurity for analysing malware in near real-time, thus eliminating the weaknesses of traditional detection methods. \\

Our approach uses NetFlow data for analysis. NetFlow records provide enough information to uniquely identify traffic using attributes such as 5-tuples and other fields, but do not expose private or personally-identifiable information (PII). NetFlow, along with its open standard version IPFIX, is already widely used for network monitoring and management. Availability of NetFlow data along with the privacy features makes it an efficient choice. \\

This paper is structured as follows: In Section 2, we review and present some Related Work where we discuss relevant topics on NetFlow, Machine Learning, Detection and Clustering Methods. Section 3 clarifies the objective of the project and the methodology used. In Section 4, we present the detail of our Data Analysis, explaining the chosen dataset and the steps involved in feature extraction and feature selection. We then present the Results in Section 5, including the metrics and algorithms used. And finally, we present our Conclusions in Section 6.

\section{Review of background and related work}
\subsection{NetFlow}
NetFlow is a feature developed by Cisco that characterizes network operations (\cite{claise2004cisco}). Network devices (routers and switches) can be used to collect IP traffic information on any of its interfaces where NetFlow is enabled. This information, known as Traffic Flows, is then collected and analyzed by a central collector. \\

NetFlow has since become an industry standard (\cite{claise2013rfc}) for capturing session data. NetFlow data  provides information that can be used to (1) identify network traffic usage and status of resources, and (2) detect network anomalies and potential attacks. It can assist in identifying devices that are hogging bandwidth, finding bottlenecks, and improving network efficiency. Tools such as NfSen/NfDump\footnote{\href{https://github.com/phaag/nfdump}{https://github.com/phaag/nfdump}} can analyze NetFlow data and track traffic patterns, which is useful for network monitoring and management. There is also an increasing number of threat analytics and anomaly detection tools that use NetFlow traffic. This paper focuses on extracting NetFlow information for forensics purposes. \\

NetFlow v5, NetFlow v9, and the open standard IPFIX are widely used. NetFlow v5 records include the IP 5-tuple data, documenting the source and destination IP addresses, source and destination ports, and the transport protocol involved in each IP flow. NetFlow v9 and IPFIX are extensible, which allow them to include other fields such as usernames, MAC addresses, and URLs.

\subsubsection{Traffic Flow data}
Flows are defined as unidirectional sequence of packets with some common properties passing through a network device (\cite{claise2004rfc}). Flow records include information such as IP addresses, packet and byte counts, timestamp, Type of Service (ToS), application ports, input and output interfaces, among others. The 5-tuple consisting of client IP, client port number, server IP, server port number, and protocol included in the flow data is important for identifying connection. By investigating the correlation between flows, we can find meaningful traffic patterns and node behaviours. \\

The output NetFlow data has the following attributes:

\begin{itemize}
    \item StartTime: the start time of the recorded flow
    \item Dur: duration of the flow
    \item Proto: protocol used (TCP, UDP, etc)
    \item SrcAddr: Source IP address
    \item Sport: Source Port
    \item Dir: Direction of communication
    \item DstAddr: Destination Address
    \item Dport: Destination Port
    \item State: the Protocol state
    \item sTos: source Type of Service
    \item dTos: destination Type of Service
    \item TotPkts: total number of packets exchanged 
    \item TotBytes: total bytes exchanged
    \item SrcBytes: number of bytes sent by source
    \item Label: label assigned to this netflow

\end{itemize}
\subsubsection{NetFlow for Anomaly Detection}
Traditional anomaly detection methods such as intrusion detection and deep packet inspection (DPI) generally require raw data or signatures published by manufacturers. DPI provides more accurate data, but it is also more computationally expensive. DPI does not work with encrypted data. It is also subject to privacy concerns because it includes user sensitive information. With the trends gearing towards data privacy and encryption, raw data may also not be easily available (\cite{kozik2017pattern}). \\

Netflow data, on the other hand, does not contain such sensitive information and is widely used by network operators. With proper analysis methods, NetFlow data can be an abundant source of information for anomaly detection. One major drawback to NetFlow has to do with the volume of data generated, particularly for links with high load. This could have an effect on the accuracy of the results and the amount of processing to be done. A lot of relevant work have already been performed to evaluate and process NetFlow data to address this. One method is by looking at NetFlow sampling (\cite{wagner2011machine}).

\subsection{Machine Learning}
Machine learning is a data analytics tool used to effectively perform specific tasks without using explicit instructions, instead relying on patterns and inference. (\cite{ML}). Machine learning capabilities are utilized by many applications to solve network and security-related issues. It can help project traffic trends and spot anomalies in network behaviour. Large providers have embraced machine learning and incorporated in tools and cloud-based intelligence systems to identify malicious and legitimate content, isolate infected hosts, and provide an overall view of the network's health (\cite{ML_Cisco}). \\

\subsubsection{Machine Learning for Botnet Detection}
The application of Machine Learning for botnet detection has been widely researched. \cite{stevanovic2014efficient} developed a flow-based botnet detection system using supervised machine learning. \cite{santana2018we} explored a couple of Machine Learning models to characterize their capabilities, performance and limitations for botnet attacks. \\

Machine learning has also been seen as a solution for evaluating NetFlows or IP-related data, where the main issue would be selecting parameters that could achieve high quality of results (\cite{wagner2011machine}). 

\subsection{Machine Learning Methods}
Several papers on NetFlow-based detection have used a number of Machine Learning techniques. \cite{kozik2018cost} presented distributed ELM, Random Forest, and Gradient-Boosted Trees as cost-sensitive approaches for Cybersecurity. In \cite{fruehwirt2014using}, these approaches are used to gain better results and flexibility. A technique called classification voting, based on decision trees and NaiveBayes, was used because it was shown to achieve high accuracy. \cite{hou2018machine} investigated DDoS tools using C.45 decision tree, Adaboost, and Random Forest algorithms. \cite{stevanovic2014efficient} analyzed and compared Random Forest and Multi-Layer Perceptron. In \cite{wagner2011machine}, the authors used Support Vector Machines (SVM) to detect and classify benign traffic from attacks. \\
 
Some of the common Machine Learning methods are described below.

\subsubsection{Extreme Learning Machine}
Extreme Learning Machine (ELM) is a learning algorithm that utilizes feedforward neural networks with a single layer or multiple layers of hidden nodes. These hidden nodes are tuned at random and their corresponding output weights are analytically determined by the algorithm. According to the creators, this learning algorithm can produce good generalization performance and can learn a thousand times faster than conventional learning algorithms for feedforward neural networks (\cite{huang_elm}).

\begin{figure}[H]
\centering
	\includegraphics[width=0.5\textwidth]{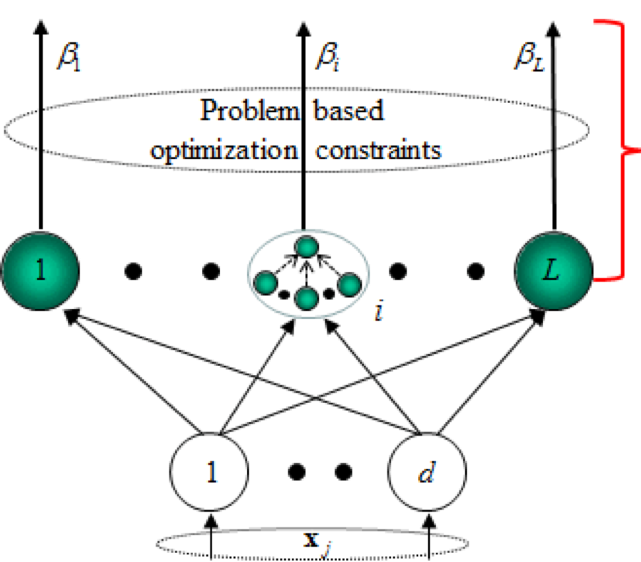}
    \caption[figure]{Simplified illustration of ELM Algorithm} (\cite{huang2015extreme})
    \label{elm}
\end{figure}

\subsubsection{Random Forest}
Random Forest (RF) is a supervised machine learning algorithm that involves the use of multiple decision trees in order to perform classification and regression tasks (\cite{ho_randomforest}). The Random Forest algorithm is considered to be an ensemble machine learning algorithm as it involves the concept of majority voting of multiple trees. The algorithm's output, represented as a class prediction, is determined from the aggregate result of all the classes predicted by the individual trees. Recent studies have explored the capabilities of Random Forest in security attacks, specifically in injection attacks, spam filtering, malware detection and more (\cite{kapoor_recommendersystems} and \cite{khorshidpour_randomforest}).

\begin{figure}[H]
\centering
	\includegraphics[trim={52 90 52 67}, clip, width=0.5\textwidth]{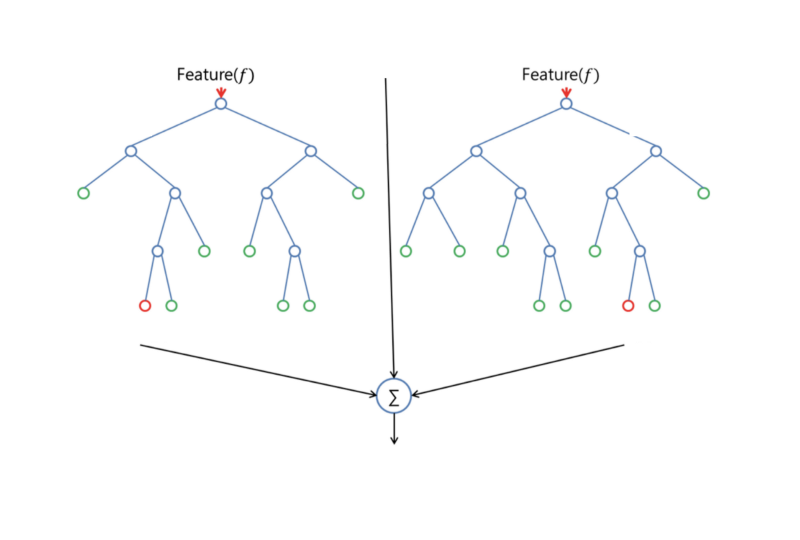}
    \caption[figure]{Simplified illustration of the Random Forest Algorithm}
    (\cite{khorshidpour_randomforest})
    \label{random_forest}
\end{figure}

\subsubsection{Gradient Boosting} 
Gradient boosting is a machine learning technique for regression and classification problems, which produces a prediction model in the form of an ensemble of weak prediction models, typically decision trees (\cite{ML}). It combines the elements of a loss function, weak learner, and an additive model. The model adds weak learners to minimize the loss function. The basic assumption with gradient boosting is to repetitively leverage the patterns in residuals and strengthen a model with weak predictions and make it better (\cite{grover2017gradient}). When it reaches a stage wherein the residuals do not have any pattern that could be modeled, then modeling of residuals will be stopped (otherwise it might lead to overfitting). Mathematically, this means minimizing the loss function such that test loss reaches its minima.

\begin{figure}[H]
\centering
	\includegraphics[width=0.5\textwidth]{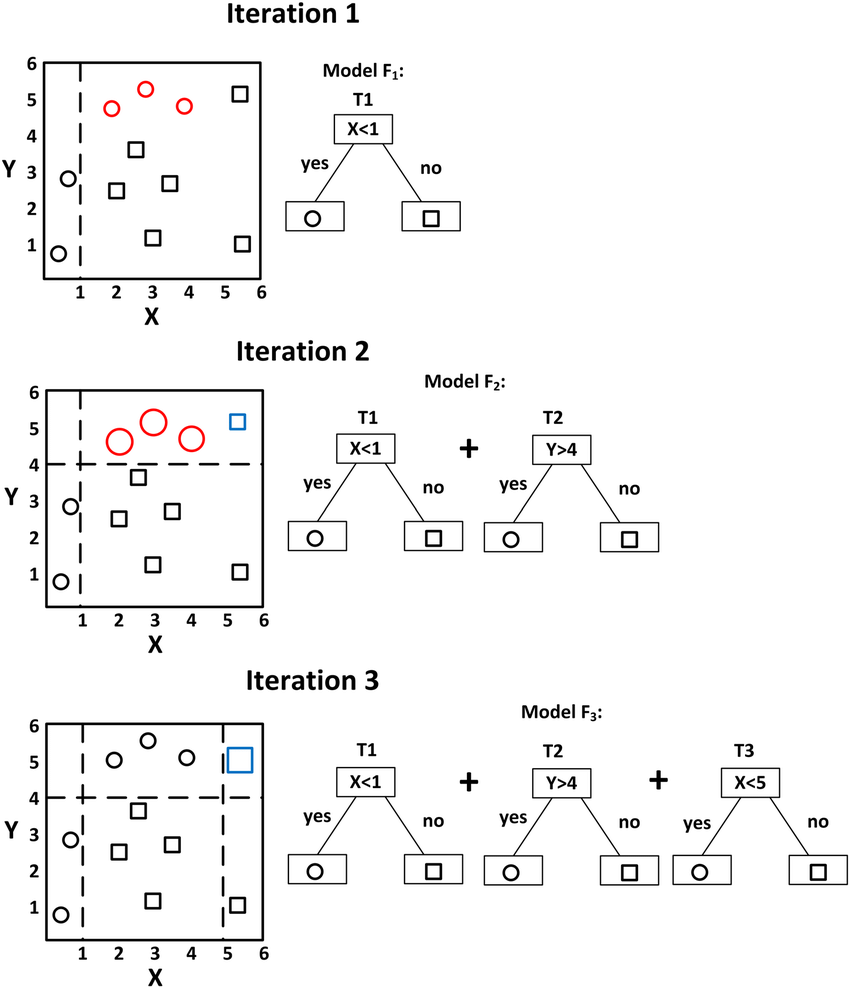}
    \caption[figure]{Simplified illustration of the Gradient Boosting Algorithm}
    (\cite{grover2017gradient})
    \label{gradient-boosting}
\end{figure}

\subsubsection{Support Vector Machine}
Support Vector Machine (SVM) is a supervised learning model used for regression and classification analysis (\cite{svm}). It is highly preferred for its high accuracy with less computation power and complexity. SVM is also used in computer security, where they are used for intrusion detection. For example, One class SVM was used for analyzing records based on a new kernel function (\cite{wagner2011machine}) and accurate Internet traffic classification (\cite{yuan2010svm}).

\begin{figure}[H]
\centering
	\includegraphics[width=0.5\textwidth]{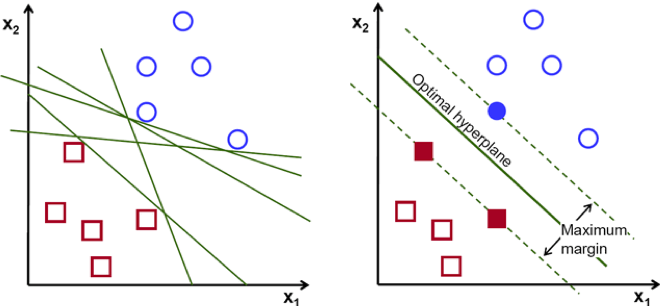}
    \caption[figure]{Simplified illustration of the Support Vector Machine}
    (\cite{introsvm})
    \label{svm}
\end{figure}

\subsubsection{Logistic Regression}
Logistic regression is a supervised learning model that is used as a method for binary classification. The term itself is borrowed from Statistics. At the core of the method, it uses logistic functions, a sigmoid curve that is useful for a range of fields including neural networks. Logistic regression models the probability for classification problems with two possible outcomes. Logistic regression can be used to identify network traffic as malicious or not (\cite{bapat2018logregress}).

\begin{figure}[H]
\centering
	\includegraphics[width=0.5\textwidth]{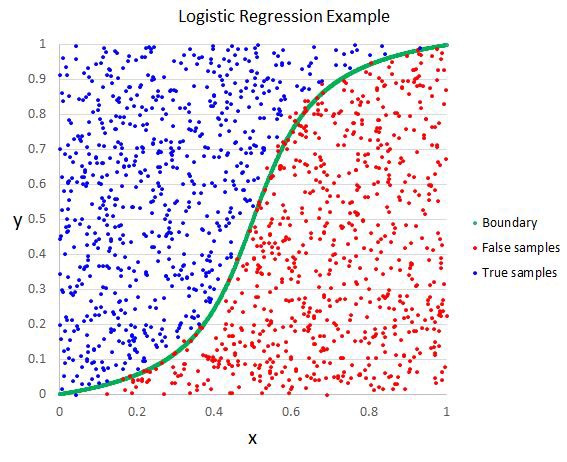}
    \caption[figure]{Simplified Illustration of the Support Vector Machine}
    (\cite{regression})
    \label{logistic-regression}
\end{figure}

\subsection{Anomaly Detection Methods}
Anomalies are objects or incidents that deviate from the normal. Therefore, anomaly detection refers to the identification of these anomalies or rare items, events or observations which raise suspicions by differing significantly from majority of the data (\cite{anomaly}). \\

In Machine Learning, anomaly detection are applied in a variety of fields including intrusion detection, fraud detection, and detecting ecosystem disturbances. There are three broad categories of anomaly detection: unsupervised, supervised, and semi-supervised. Some of the popular detection techniques include density-based k-nearest neighbour, one-class SVM, Bayesian Networks, Cluster-analysis-based outlier detection among others. A number of analysis systems use the above detection techniques. Here we discuss these three systems - CAMNEP, MINDS and Xu - which was compared in detail in \cite{botnet}. 

\subsubsection{CAMNEP}
The Cooperative Adaptive Mechanism for Network Protection (CAMNEP) is a network intrusion detection (\cite{rehak2008camnep}) system. The CAMNEP system uses a set of anomaly detection model that maintain a model of expected traffic on the network and compare it with real traffic to identify the discrepancies that are identified as possible attacks. It has three principal layers that evaluate the traffic: anomaly detectors, trust models, and anomaly aggregators. The anomaly detector layer analyzes the NetFlows using various anomaly detection algorithms, each of which uses a different set of features. The output are aggregated into events and sent into the trust models. The trust model maps the NetFlows into traffic clusters. NetFlows with similar behavioural patterns are clustered together. The aggregator layer creates the composite output that integrates the individual opinion of several anomaly detectors.  

\subsubsection{MINDS}
The Minnesota Intrusion Detection System (MINDS) uses a suite of data mining techniques to automatically detect attacks (\cite{minds}). It builds a context information for each evaluated NetFlow using the following features: the number of NetFlows from the same source IP address as the evaluated NetFlow, the number of NetFlows toward the same destination host, the number of NetFlows towards the same destination host from the same source port, and the number of NetFlows from the same source host towards the same destination port. The anomaly value for a NetFlow is based on its distance to the normal sample (\cite{rehak2008camnep}). 

\subsubsection{Xu}
This algorithm was proposed by Xu et al. The context of each NetFlow to be evaluated is created with all the NetFlows coming from the same source IP address. The anomalies are detected by some classification rules that divide the traffic into normal and anomalous flows. 

\subsection{Clustering Methods based on Botnet behavior}
\subsubsection{BClus}
The BClus Method is a an approach that uses behavioural-based botnet detection.  BClus creates a model of known botnet behaviours and uses them to detect similar traffic on the network. The purpose of the method is to cluster the traffic sent by each IP address and to recognize which clusters have a behaviour similar to the botnet traffic. (\cite{rehak2008camnep}). 

\subsubsection{K-Means Algorithm}
K-means algorithm is an unsupervised machine learning algorithm popular for clustering and data mining. It works by storing k centroids that is used to define a cluster. In \cite{terzi2017big}, the aggregated NetFlows are clustered based on the K-means algorithm, and clusters are expected to occur based on traffic behaviour.

\subsection{Malware Infection Process Stage Detection}
\subsubsection{BotHunter}
The BotHunter method is useful for detection of infections and for coordination dialog of botnets. It is done by matching state-based infection sequence models. It consists of a correlation engine that aims at detecting specific stages of the malware infection process (\cite{rehak2008camnep}). It uses an adaptive version of the Snort IDS with two proprietary plugins, called Statistical Scan Anomaly Detection Engine (SCADE) and Statistical Payload Anomaly Detection Engine (SLADE).

\subsubsection{B-ELLA}
The Balanced Efficient Lifelong Learning (B-ELLA) framework is another approach to cyber attack detection (\cite{bella}). It is an extension of the Efficient Lifelong Learning (ELLA) framework that copes with the problem of data imbalance. The original ELLA framework allows for building and maintaining a sparsely shared basis for task model (or classifiers). \\ 

Lifelong Learning (LL)\footnote{\href{http://lifelongml.org/}{http://lifelongml.org/}} in general is an advanced Machine Learning paradigm that focuses on learning and accumulating knowledge continuously, and uses this knowledge to adapt or help future learning (\cite{lifelonglearning}). B-ELLA is a practical application of Lifelong Learning in the field of cybersecurity.

\section{Aim of the Project}
\subsection{Objectives}
The main objective of this project is to detect malware or botnet traffic from a NetFlow dataset using different Machine Learning approaches. \\

More specifically, our proposed approach seeks to:
\begin{itemize}
\item Detect malware or botnet traffic from a Netflow data. The system should take any Netflow dataset of any size, clean or with malware, and classify as either normal or attack traffic.
\item Compare a variety of Machine Learning methods and recommend the suitable one for specific use cases.
\end{itemize}

\subsection{Methodology}
To achieve the above objectives, we follow the methodology as described below.

\begin{enumerate}
    \item \textbf{Selecting a Dataset} The first part of the methodology is collecting traffic flow data. We can do this by sourcing actual data traffic from a known organization and extracting NetFlows. In the absence of actual data traffic, another option is to use a collection of public domain datasets. Well-known datasets for this purpose are CTU-13 (\cite{CTU13}), KDDCUP99\footnote{\href{http://kdd.ics.uci.edu/databases/kddcup99/kddcup99.html}{http://kdd.ics.uci.edu/databases/kddcup99/kddcup99.html}} and CIC-IDS-2017\footnote{\href{https://www.unb.ca/cic/datasets/ids-2017.html}{https://www.unb.ca/cic/datasets/ids-2017.html}}. We have chosen to use CTU-13 over other public datasets because it is highly available and has been used quite extensively for many similar research studies in the past. We discuss more detail about the dataset in Section \ref{ctu13_dataset_section}.
    We then examine the dataset statistically to identify common features and frequencies. The features are the raw attributes in the Netflow data - StartTime, Duration, Proto, SrcAddr, Sport, Dir, DstAddr, Dport, State, sTos, dTos, TotPkts, TotBytes, SrcBytes and Label. This group of features characterizes the flows.
    \item \textbf{Feature Extraction} Once we have selected a dataset, we then identify and extract the features. This step is a very important part of the methodology. NetFlow data contains categorical features that have to be encoded into numerical or boolean values, which would result in a matrix size that is too big and cause memory issues. To trim down the amount of data to be processed, we use a schema to sample the NetFlow traffic using time window. We have chosen a time window of 2 minutes with a stride of 1 minute. We then preprocess the bidirectional NetFlow dataset and extract categorical and numerical characteristics that describe the dataset within the given time window.
    \item \textbf{Feature Selection} This step is required to select features from the extracted ones. It involves the use of feature selection techniques to reduce the dimension of the input training matrix. Filtering features through Pearson Correlation, wrapper methods using Backward Feature Elimination, embedded methods within the Random Forest Classifier, Principal Component Analysis (PCA), and the t-distributed Stochastic Neighbour Embedding (t-SNE) are the different techniques used for this purpose.
    \item \textbf{Comparison of Algorithms} We wanted to compare five chosen algorithms. The various Machine Learning Models to be trained in this step are: Logistic Regression, Support Vector Machine (SVM), Random Forest Classifier, Gradient Boosting, and Dense Neural Network.
    \item \textbf{Botnet Detection} The last step in our methodology involves testing our model to see if it can successfully detect botnet traffic from the CTU-13 dataset. The overall performance of botnet detection is determined from the $f_1$ score of the aforementioned models.
\end{enumerate}

\subsection{Main issue with Network Security Data}
Working with Network Security Data brings lots of challenges. First, the data are very imbalance because most of the traffic is harmless and only a tiny part of it is malicious. That causes the model to difficultly learn what is harmful. Moreover, the risk of overfitting during the training process is high because the structure of the network influences the way the model learns while a network-independent algorithm is wanted.\\

Furthermore, traffic analysis deals with a network which is a dynamic structure: communications are time dependent and links between servers may appear and disappear with new requests and new users in the network. That is why detecting new unknown botnets is a real challenge in network security.\\

To cope with all these challenges, an accurate data analysis is needed and mechanisms to prevent overfitting like cross-validation are necessary.

\newpage
\section{Data analysis}
\subsection{CTU-13 Dataset}
\label{ctu13_dataset_section}
The CTU-13 Dataset is a labeled dataset used in the literature to train botnet detection algorithms (see for example \cite{botnet} and \cite{bella}). It was created by the \cite{CTU13} and captures real botnet traffic mixed with normal traffic and background traffic. In this section, we will describe the dataset in order to find relevant features to extract, and train our models.\\

\begin{figure}[H]
\centering
	\includegraphics[width=1.0\textwidth]{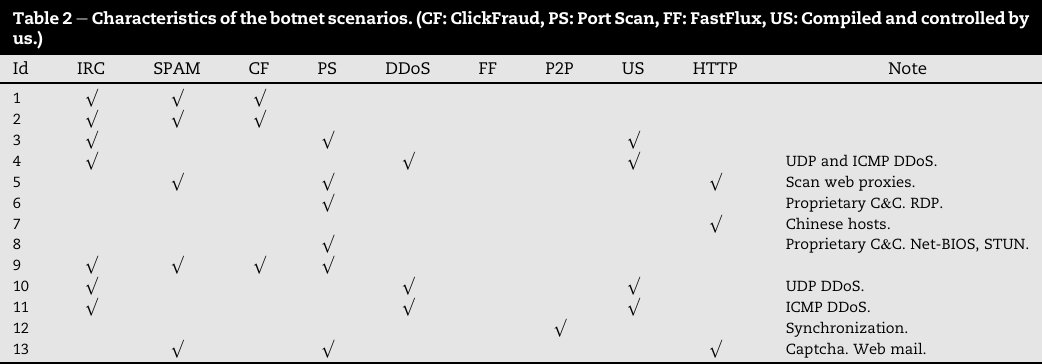}
    \caption[figure]{Scenarios of the CTU-13 Dataset (\cite{CTU13})}
    \label{ctu13}
\end{figure}

The CTU-13 Dataset is made of 13 captures of different botnet samples summarized in 13 bidirectional NetFlow\footnote{``NetFlow is a feature that was introduced on Cisco routers around 1996 that provides the ability to collect IP network traffic as it enters or exits an interface.'' \cite{netflow}} files (see Figure \ref{ctu13}).

We focus here our presentation on the first scenario, which uses a malware called \textit{Neris}. The capture lasted 6.15 hours during which the botnet used HTTP based C\&C channels\footnote{``A Command-and-Control [C\&C] server is a computer controlled by an attacker or cybercriminal which is used to send commands to systems compromised by malware and receive stolen data from a target network.'' \cite{candc}} to send SPAM and perform Click-Fraud. The NetFlow file weights 368 Mo and is made of 2 824 636 bidirectional communications described by 15 features:
\begin{enumerate}
    \item \textbf{StartTime}: Date of the flow start (from 2011/08/10 09:46:53 to 2011/08/10 15:54:07)
    \begin{figure}[H]
    \centering
    	\includegraphics[trim={11 13 12 13}, clip, width=0.68\textwidth]{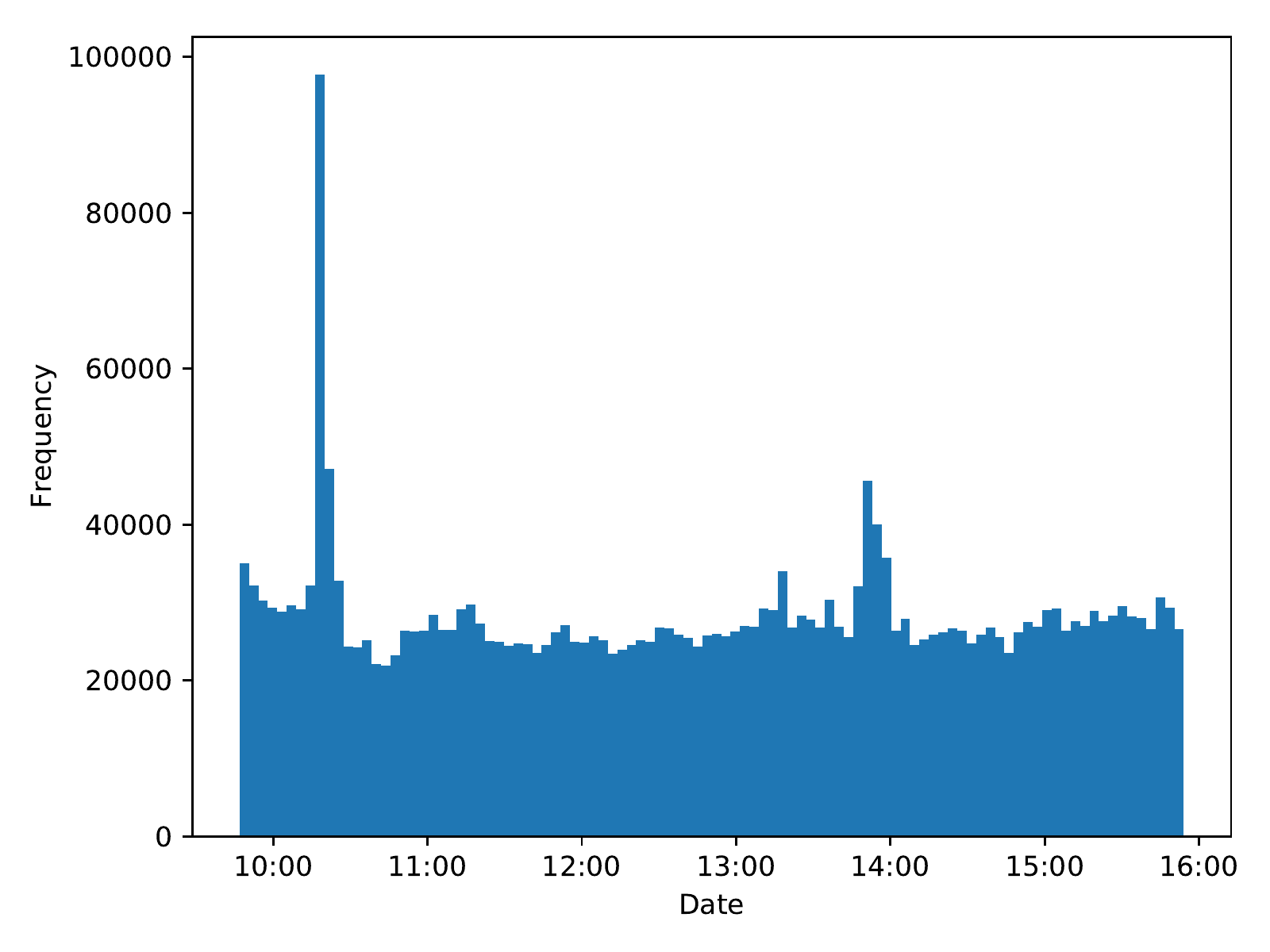}
        \caption[figure]{Frequency of NetFlows wrt Time}
        \label{starttime}
    \end{figure}
    As shown in Figure \ref{starttime}, the start times of NetFlows seem uniformly distributed during the experiment with an exception at around 10:20 with a pic of starting NetFlows.
    \item \textbf{Dur}: Duration of the communication in seconds\\
    (min: 0 s.; max: 3 600 s.; mean: 432 s.; std: 996 s.; median: 0.001 s.; 3\textsuperscript{rd} quantile: 9 s.)
    \begin{figure}[H]
    \centering
    	\includegraphics[trim={11 13 12 13}, clip, width=0.68\textwidth]{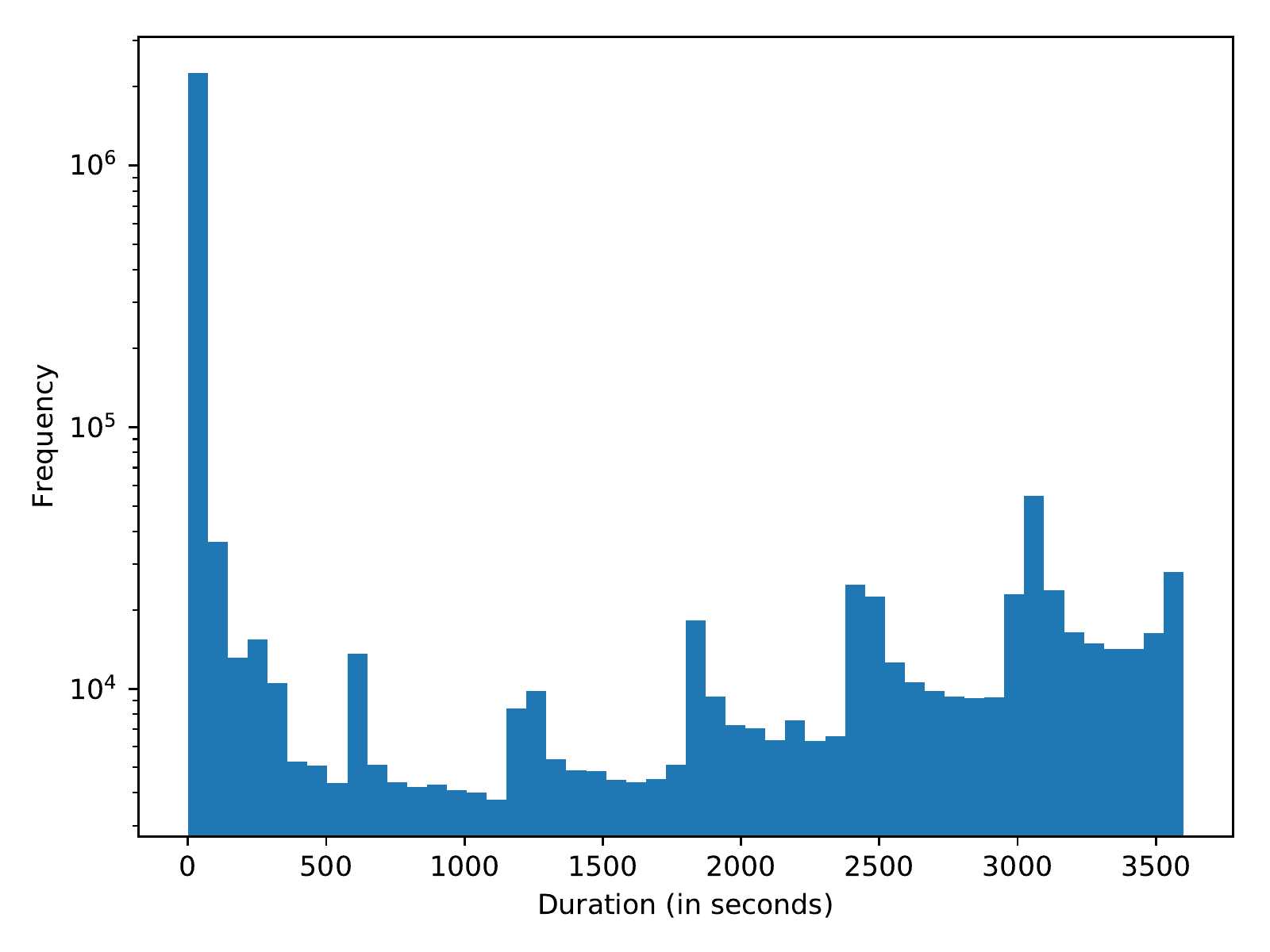}
        \caption[figure]{Duration of the communications}
        \label{dur}
    \end{figure}
    Figure \ref{dur} shows that the majority of the communications last for a very short time (less than 1 ms. for half of them). One can also notice frequency peaks each time a minute is reached.
    \item \textbf{Proto}: Transport Protocol used (15 protocols in total)
    \begin{figure}[H]
    \centering
    	\includegraphics[trim={30 30 30 30}, clip, width=0.72\textwidth]{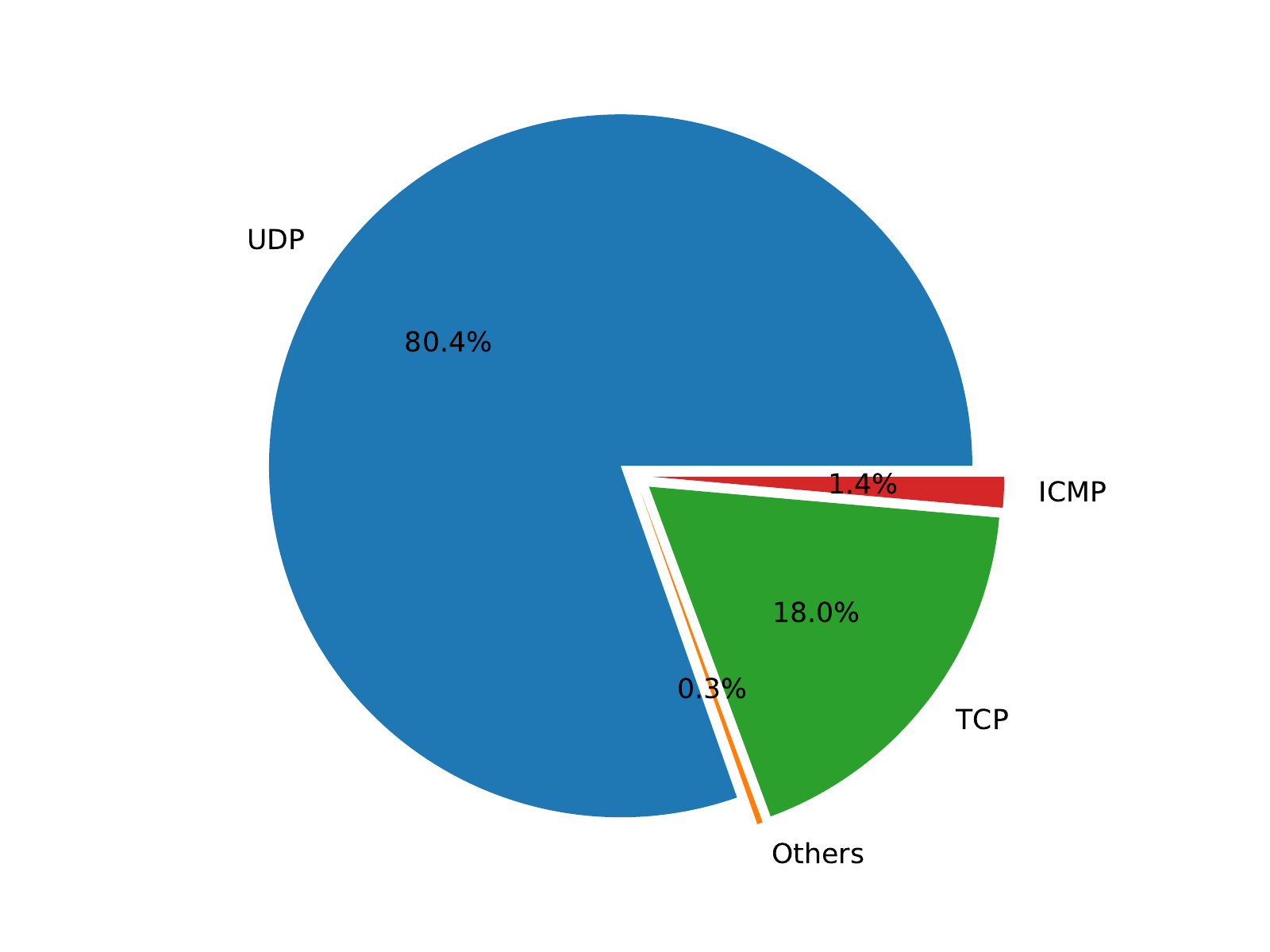}
        \caption[figure]{Distribution of the 15 Protocols}
    \end{figure}
    \item \textbf{SrcAddr}: IP address of the source (542 093 addresses in total)
    \begin{figure}[H]
    \centering
    	\includegraphics[trim={30 30 30 30}, clip, width=0.72\textwidth]{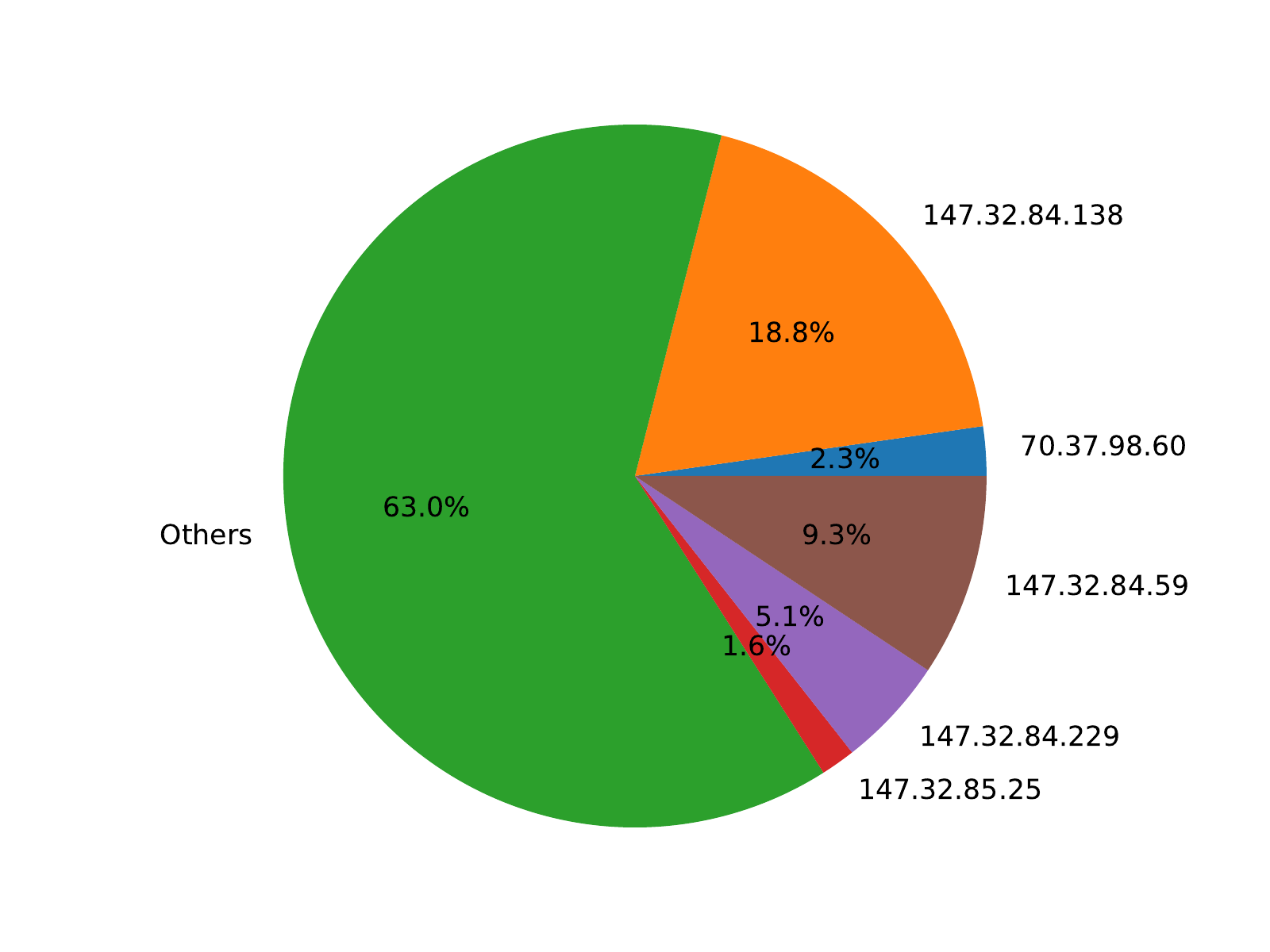}
        \caption[figure]{Distribution of the Source IP addresses}
    \end{figure}
    The huge number of different IP addresses prevents the categorization of the data from SrcAddr because of memory concerns. Another way needs to be found to represent the huge amount of data.
    \item \textbf{Sport}: Source port (64 752 ports in total)
    \begin{figure}[H]
    \centering
    	\includegraphics[trim={30 30 30 30}, clip, width=0.69\textwidth]{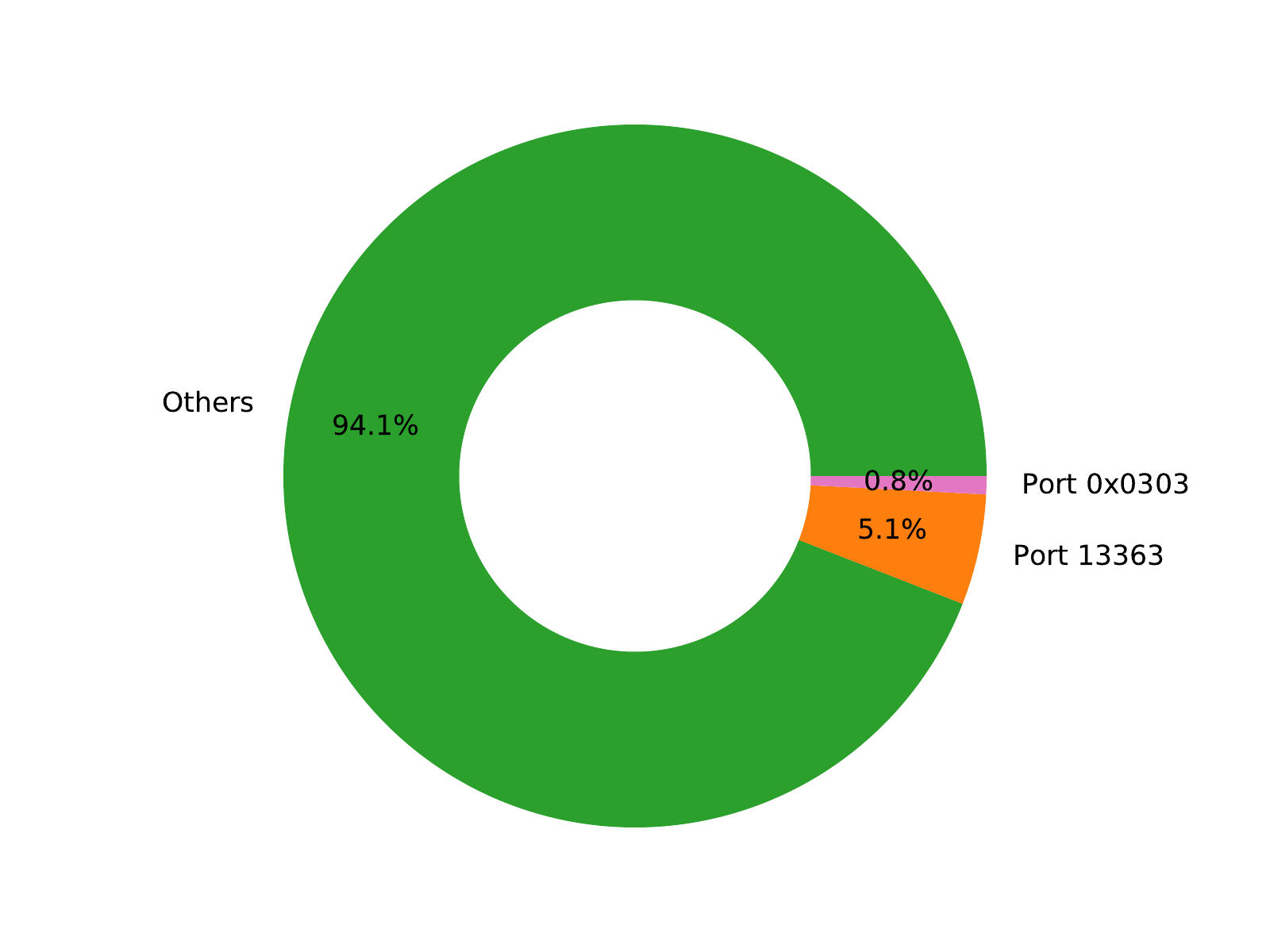}
        \caption[figure]{Distribution of the Source ports}
    \end{figure}
    \item \textbf{Dir}: Direction of the flow (78\% bidirectional, 22\% from source to destination)
    \item \textbf{DstAddr}: IP address of the destination (119 296 addresses in total)
    \begin{figure}[H]
    \centering
    	\includegraphics[trim={30 30 29 30}, clip, width=0.69\textwidth]{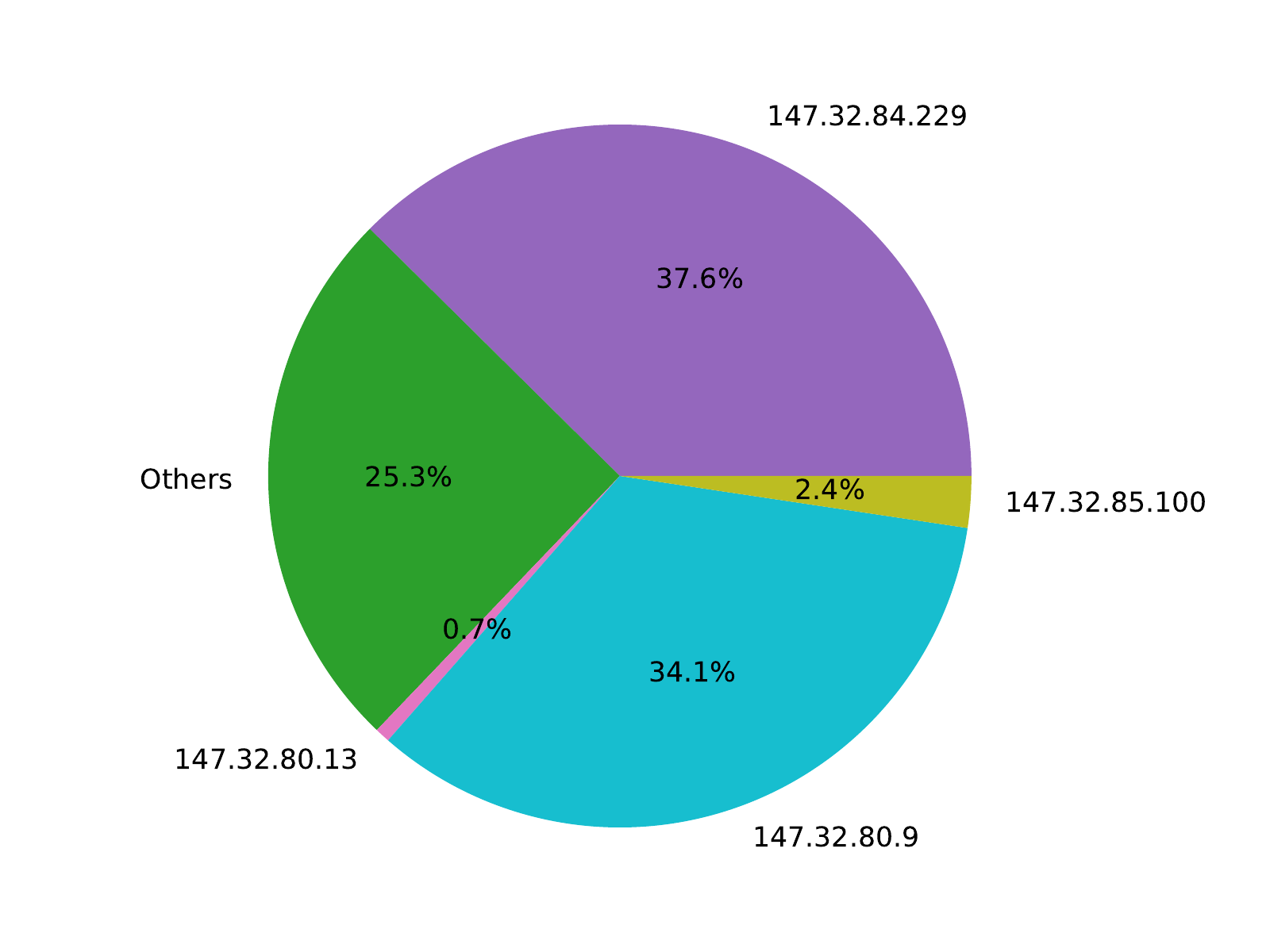}
        \caption[figure]{Distribution of the Destination IP addresses}
    \end{figure}
    \item \textbf{Dport}: Destination port\footnote{Port 53: Domain Name System (DNS); Port 80: Hypertext Transfer Protocol (HTTP); Port 443: Hypertext Transfer Protocol over TLS/SSL (HTTPS); Port 6881: BitTorrent (Unofficial)} (73 786 ports in total)
    \begin{figure}[H]
    \centering
    	\includegraphics[trim={30 30 30 30}, clip, width=0.68\textwidth]{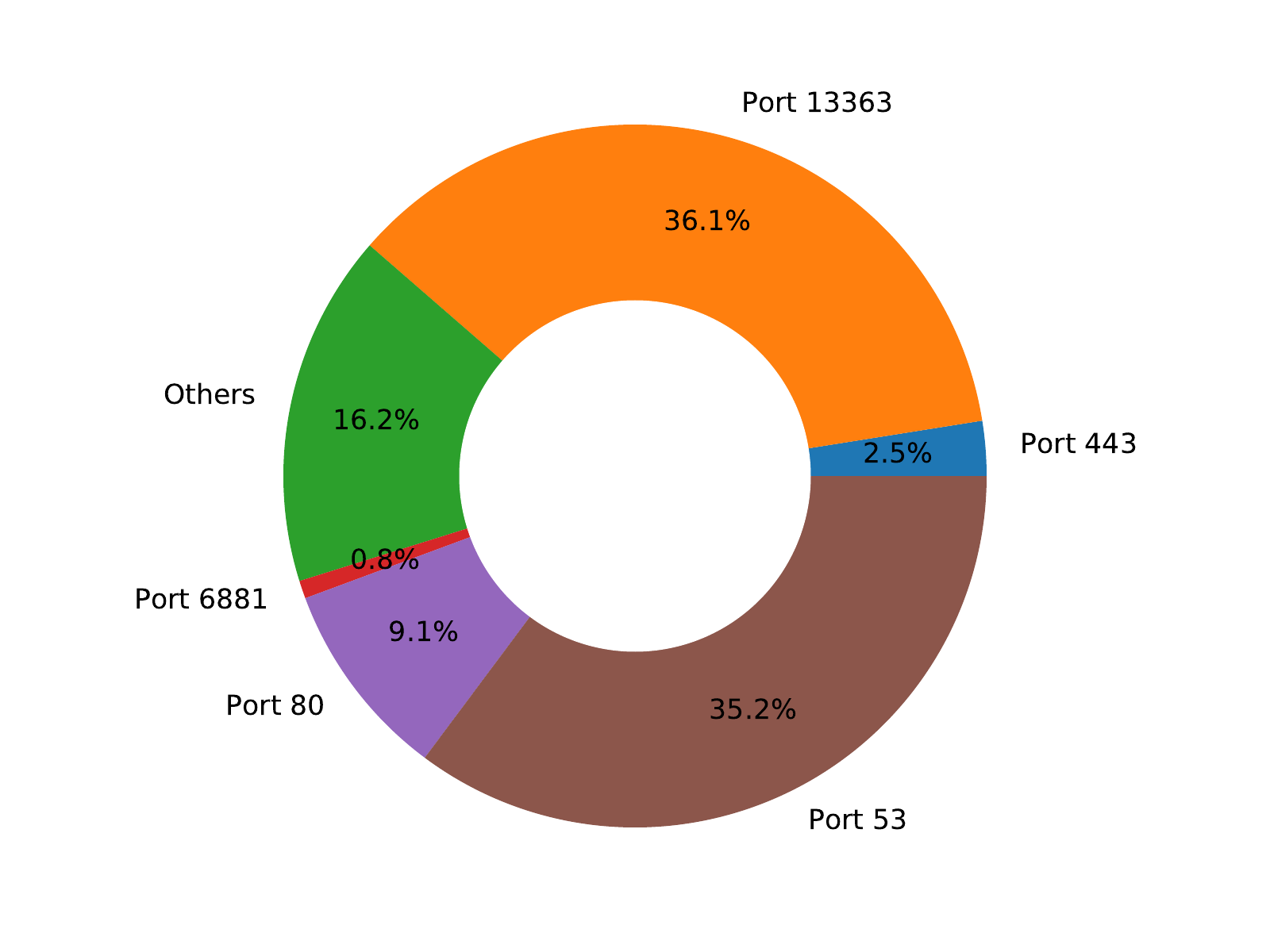}
        \caption[figure]{Distribution of the Destination ports}
    \end{figure}
    \item \textbf{State}: Transaction state\footnote{The state is protocol dependent and \_ is a separator for one end of the connection. Examples of states: CON = Connected (UDP); INT = Initial (UDP); URP = Urgent Pointer (UDP); F = FIN (TCP); S = SYN = Synchronization (TCP); P = Push (TCP); A = ACK = Acknowledgement (TCP); R = Reset (TCP); FSPA = All flags - FIN, SYN, PUSH, ACK (TCP)} (230 states in total)
    \begin{figure}[H]
    \centering
    	\includegraphics[trim={30 30 30 30}, clip, width=0.68\textwidth]{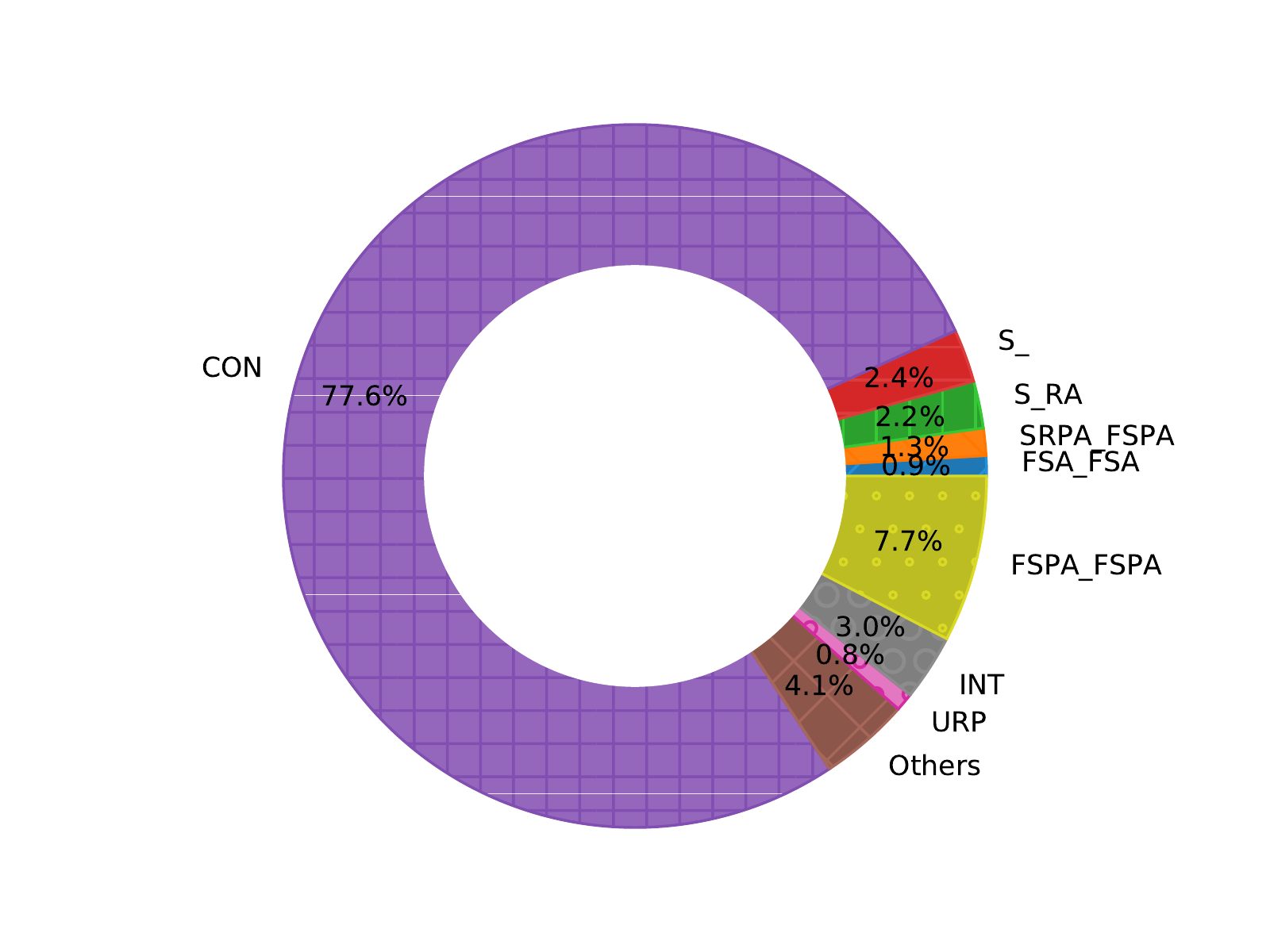}
        \caption[figure]{Distribution of Transaction states}
    \end{figure}
    \item \textbf{sTos}: Source TOS\footnote{Depict the priority of the packet (0 or 192: Routine; 1: Priority; 2: Immediate; 3: Flash; ...)} byte value (0 for 99.9\% of the communications)
    \item \textbf{dTos}: Destination TOS byte value (0 for 99.99\% of the communications)
    \item \textbf{TotPkts}: Total number of transaction Packets (min: 1 packet; max: 2 686 731 packets)
    \begin{figure}[H]
    \centering
    	\includegraphics[trim={11 10 12 13}, clip, width=0.66\textwidth]{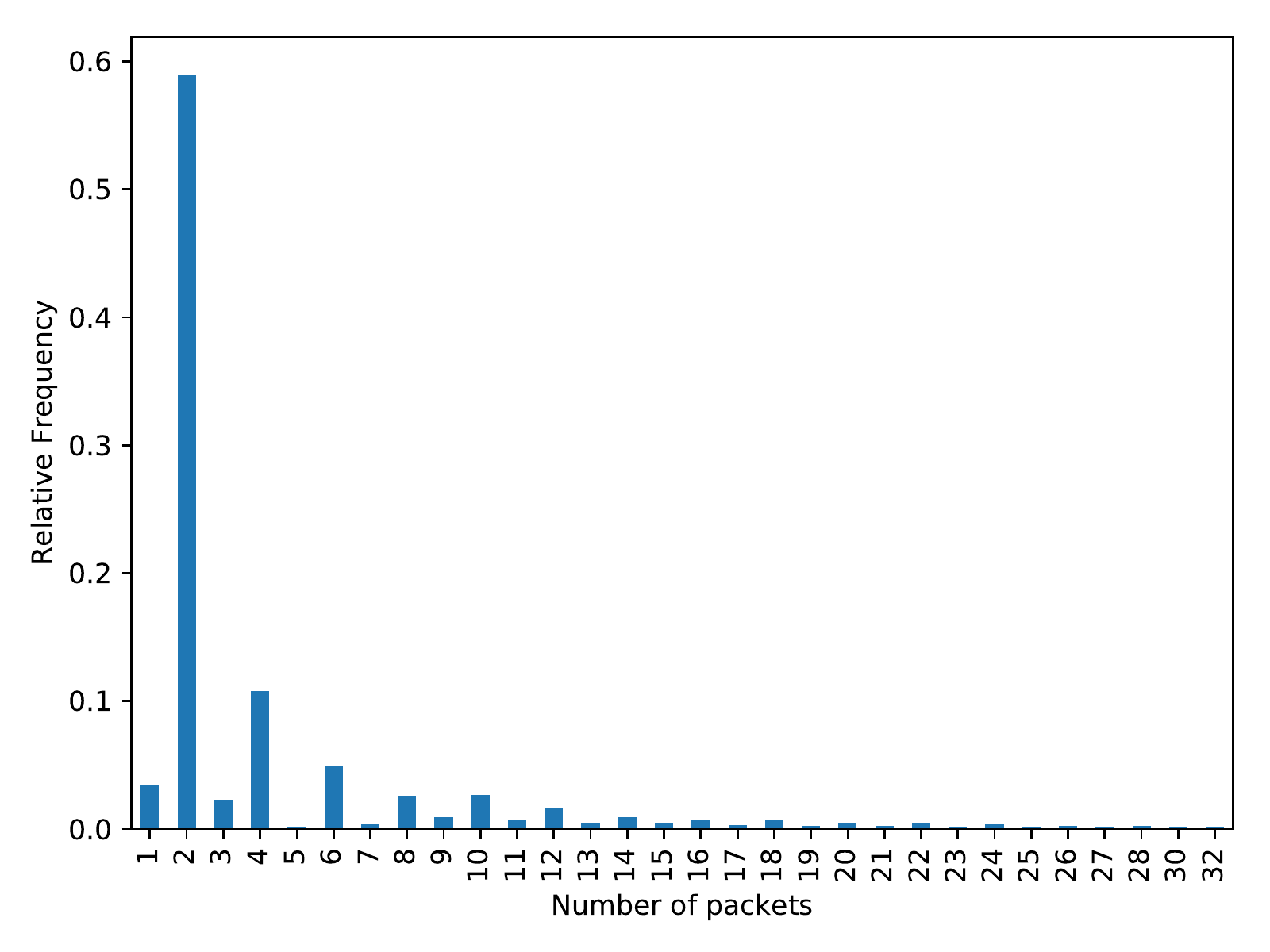}
        \caption[figure]{Most frequent Numbers of Packets}
    \end{figure}
    \item \textbf{TotBytes}: Total number of transaction Bytes (min: 60 Bytes; max: 2 689 640 464 Bytes)
    \begin{figure}[H]
    \centering
    	\includegraphics[trim={11 10 12 13}, clip, width=0.66\textwidth]{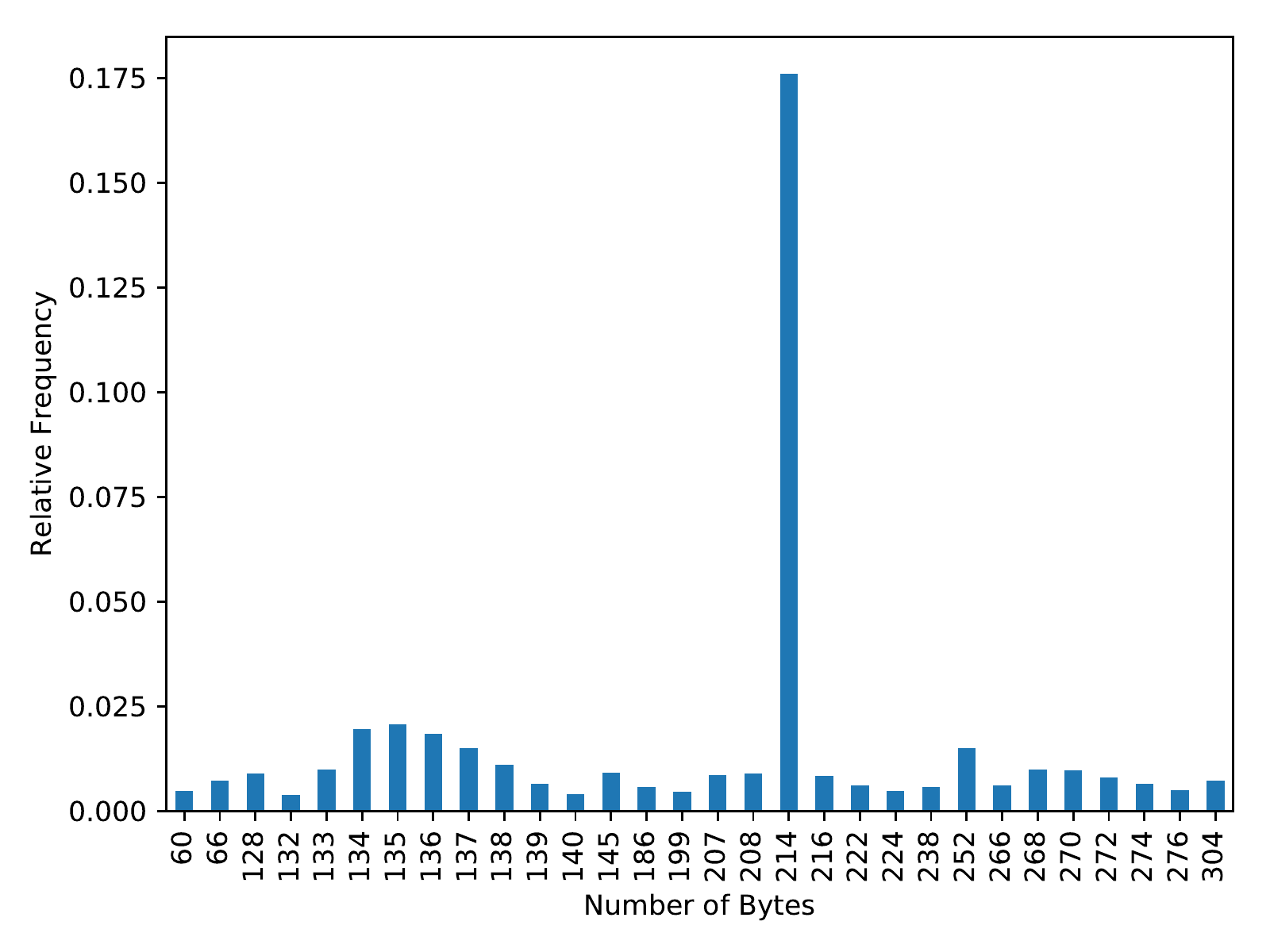}
        \caption[figure]{Most frequent Numbers of Bytes}
    \end{figure}
    \item \textbf{SrcBytes}: Total number of transaction Bytes from the Source \small{(min: 0 Byte; max: 2 635 366 235 Bytes)}
    \begin{figure}[H]
    \centering
    	\includegraphics[trim={11 10 12 13}, clip, width=0.73\textwidth]{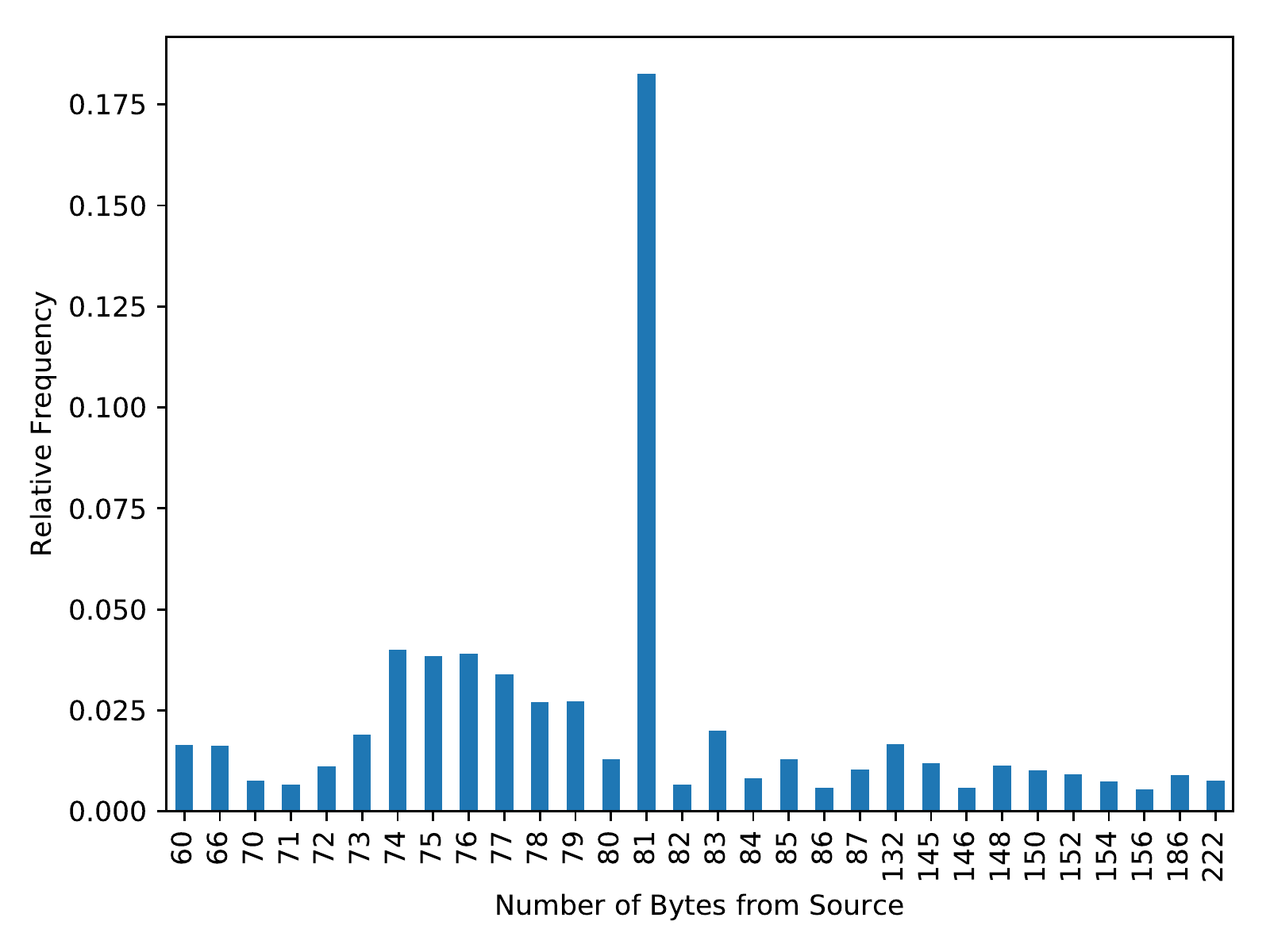}
        \caption[figure]{Most frequent Numbers of Bytes}
    \end{figure}
    \item \textbf{Label}: Label made of ``flow='' followed by a short description (Source/Destination-Malware/Application)
    \begin{figure}[H]
    \centering
    	\includegraphics[trim={11 10 12 13}, clip, width=0.73\textwidth]{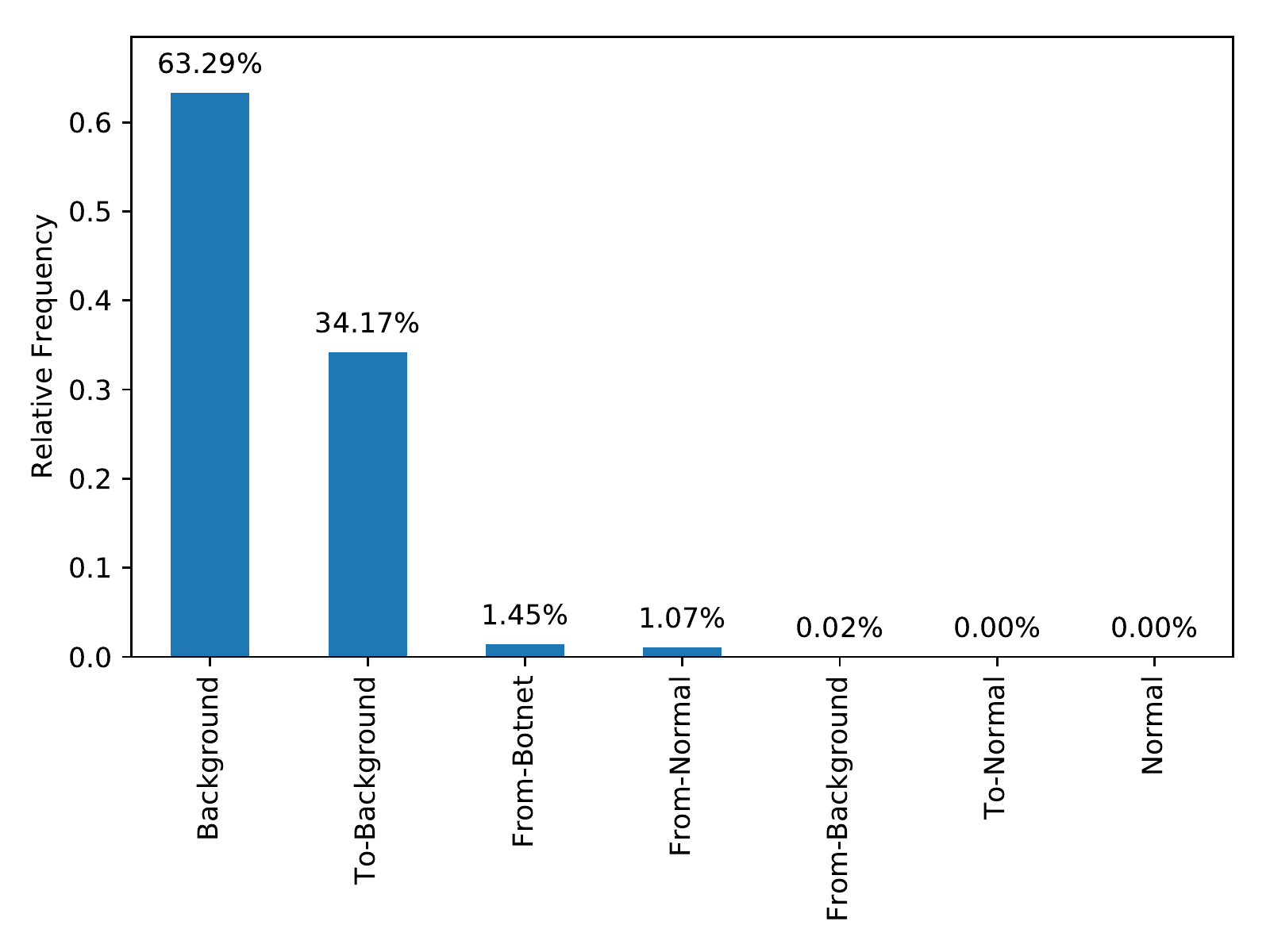}
        \caption[figure]{Distribution of the 113 Labels}
    \end{figure}
\end{enumerate}

\subsection{Feature extraction}
\label{feature_extraction}
\subsubsection{Motivation}
The main issue with NetFlow data is that most of the information is in the form of categorical features. Trying to transform the categories into boolean columns results in a boom of the matrix size causing memory errors (even after using compressed sparse matrix representation).\\

This major problem encouraged us to extract new features, based on the reviewed literature of network traffic analysis.

\subsubsection{Use of time windows}
A popular idea is to summarize data inside a time window. This is justified by the fact that botnets ``tend to have a temporal locality behavior'' (\cite{botnet}). Moreover, it enables us to reduce the amount of data and to deliver a botnet detector which gives live results after each time window (in exchange for input data details).\\

The main issue is to determine the width and the stride of the time window. The reviewed literature suggests different options (2 minutes in the BClus detection method in \cite{botnet}, 3 seconds in the MINDS algorithm in \cite{minds}, 5 minutes in the Xu algorithm in \cite{xu}) but these suggestions are often empirical. In our case, we have chosen a time window width of 2 minutes and a stride of 1 minute.\\

Another problem we encountered is that the time window gathers NetFlow communications according to the start time of the connection. Using the communication status (active or finished) resulted in a boom in the data size so we used the duration of the communication as extra features.

\subsubsection{Extracted features}
The extracted features try to represent the NetFlow communications inside a time window. To do so, the time window data are gathered by source addresses.\\

There are 2 features extracted from each categorical NetFlow characteristics (in our case: the source ports, the destination addresses and the destination ports):
\begin{itemize}
    \item the number of unique occurrences in the subgroup
    \item the normalized subgroup entropy defined as
    \begin{equation}
        E = -\sum_{x_i \in X}p(x_i)\log p(x_i)
        \text{\hspace{0.5cm}with\hspace{0.5cm}}
        p(x_i) = \frac{\#x_i}{\#X}
        \text{\hspace{0.3cm}and\hspace{0.3cm}X, the subgroup of the source address}
    \end{equation}
\end{itemize}

As for the numerical NetFlow characteristics (in our case: the duration of the communication, the total number of exchanged bytes, the number of bytes sent by the source), 5 features are extracted:
\begin{itemize}
    \item the sum
    \item the mean
    \item the standard deviation
    \item the maximum
    \item the median
\end{itemize}

All these extracted features will enable us to train different models. However, they may be redundant or irrelevant to detect botnets so a feature selection is needed.

\subsection{Feature selection}
\subsubsection{Filter and Wrapper methods}
\label{feature_selection1}
\begin{figure}[H]
\centering
    \includegraphics[width=0.73\textwidth]{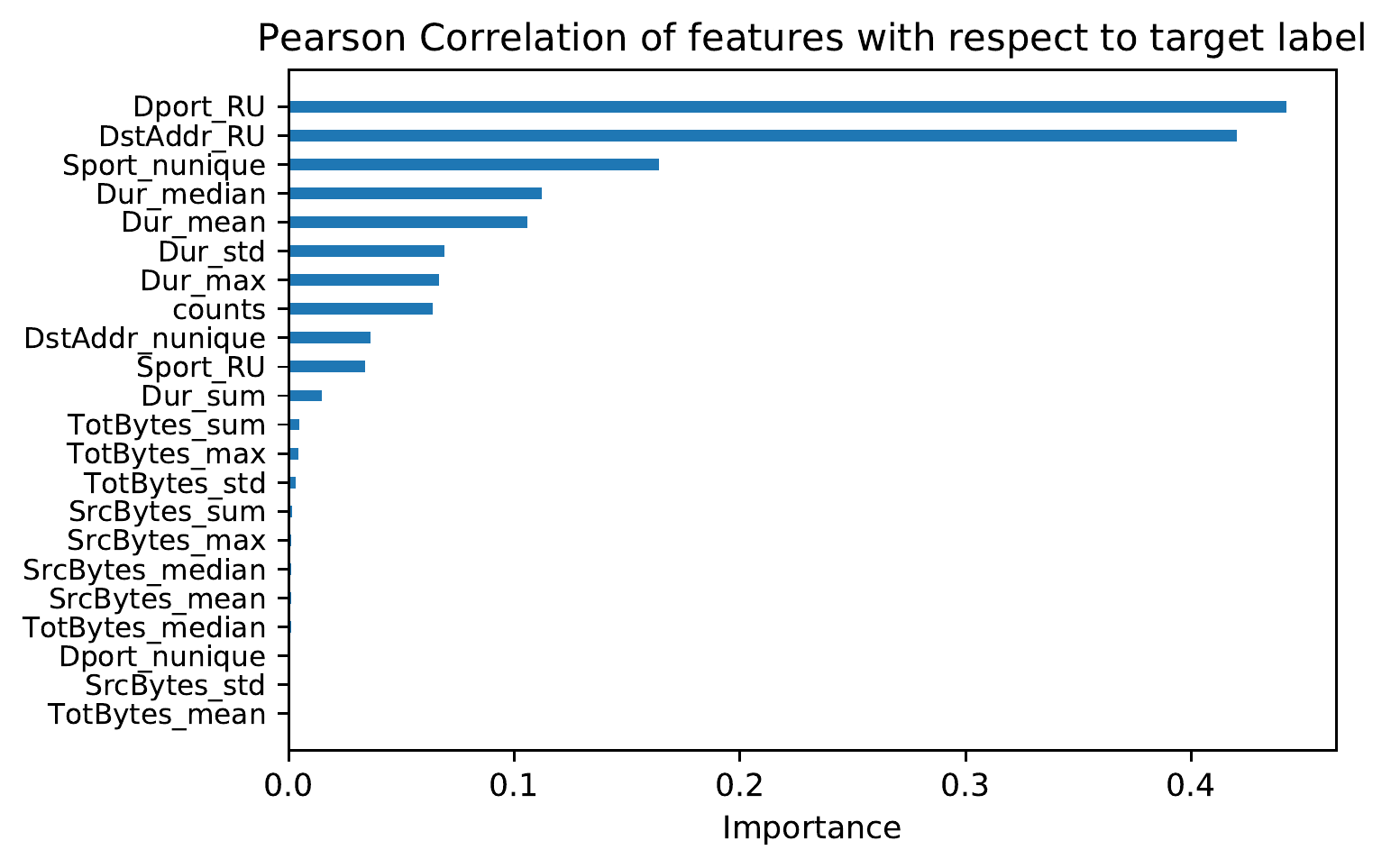}
    \caption[figure]{Pearson Correlation of features with respect to target label}
    \label{pearson_correlation}
\end{figure}

\paragraph{Filter Method}~~\\
The filter method, as the name suggests, involves filtering features in order to select the best subset of features for training the model. The best features are selected based on scores of various statistical tests that determine their correlation with the target label. The Pearson Correlation statistical test is used in our case since the pre-processed data involves continuous data. Figure \ref{pearson_correlation} illustrates the Pearson Correlation scores of the features with respect to the target label. 

Among all the features, \texttt{Dport\_RU} and \texttt{DstAddr\_RU} rank the highest, followed by the rest with significantly lower scores. From this, the best features are identified by selecting all features above a certain threshold (e.g. features with scores higher than 0.1) which yields a subset of best features containing: \texttt{Dport\_RU}, \texttt{DstAddr\_RU}, \texttt{Sport\_nunique}, \texttt{Dur\_median} and \texttt{Dur\_mean}. To conclude with the filter method, the resultant best features are compared against each other in order to remove highly correlated ones and select the most independent ones. Highly correlated features within the subset can be identified using the Feature Correlation Heatmap (Figure 20) and will be further discussed there.

\begin{figure}[H]
\centering
    \includegraphics[width=0.73\textwidth]{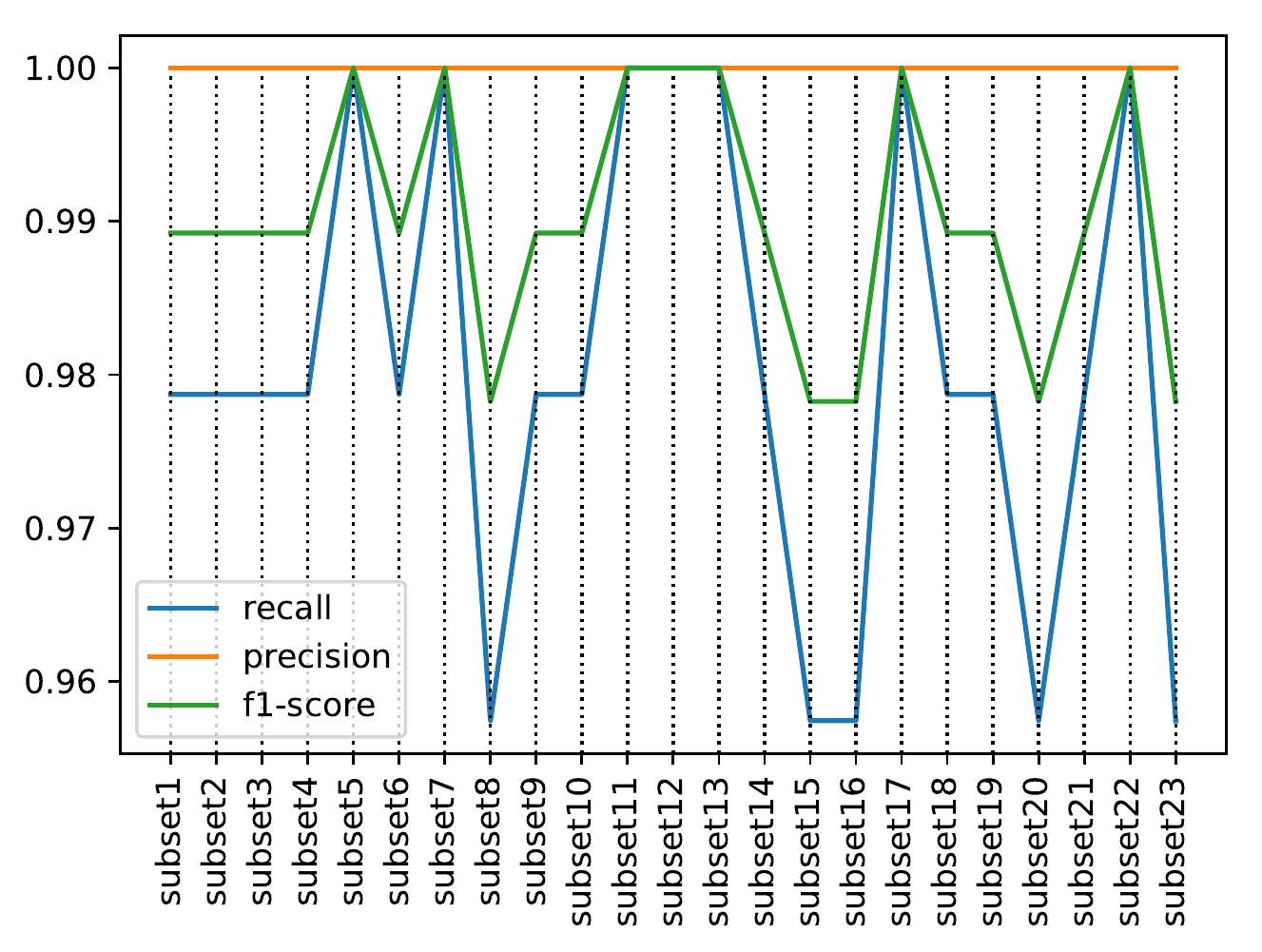}
    \caption[figure]{Backward Feature Elimination Results}
    \label{backward-elimination}
\end{figure}

\paragraph{Wrapper Method}~~\\
The wrapper method involves testing different subsets of features on a training model in an iterative fashion in order to determine the best features based on the inferences made by the training model. During each iteration, features are either added to or removed from the subset which will then be used to train the model. This is repeated until no more improvement is observed. This method is typically very computationally expensive because it needs many training iterations. There are many known wrapper methods, such as forward feature selection, backward feature elimination and exhaustive feature selection. However, the one implemented here is the backward feature elimination. It starts by training the model (Random Forest) with all of the features and removing features one by one until the $f_1$ score does not improve. Figure \ref{backward-elimination} illustrates the results using this method. 

Features removed in each subset:
    \begin{multicols}{2}
    \begin{itemize}
        \item \underline{subset 1:} No features removed
        \item \underline{subset 2:} \texttt{counts} removed
        \item \underline{subset 3:} \texttt{Sport\_nunique} removed
        \item \underline{subset 4:} \texttt{DstAddr\_nunique} removed
        \item \underline{subset 5:} \texttt{Dport\_nunique} removed
        \item \underline{subset 6:} \texttt{Dur\_sum} removed
        \item \underline{subset 7:} \texttt{Dur\_mean} removed
        \item \underline{subset 8:} \texttt{Dur\_std} removed
        \item \underline{subset 9:} \texttt{Dur\_max} removed
        \item \underline{subset 10:} \texttt{Dur\_median} removed
        \item \underline{subset 11:} \texttt{TotBytes\_sum} removed
        \item \underline{subset 12:} \texttt{TotBytes\_mean} removed
        \item \underline{subset 13:} \texttt{TotBytes\_std} removed
        \item \underline{subset 14:} \texttt{TotBytes\_max} removed
        \item \underline{subset 15:} \texttt{TotBytes\_median} removed
        \item \underline{subset 16:} \texttt{SrcBytes\_sum} removed
        \item \underline{subset 17:} \texttt{SrcBytes\_mean} removed
        \item \underline{subset 18:} \texttt{SrcBytes\_std} removed
        \item \underline{subset 19:} \texttt{SrcBytes\_max} removed
        \item \underline{subset 20:} \texttt{SrcBytes\_median} removed
        \item \underline{subset 21:} \texttt{Sport\_RU} removed
        \item \underline{subset 22:} \texttt{DstAddr\_RU} removed
        \item \underline{subset 23:} \texttt{Dport\_RU} removed
    \end{itemize}
    \end{multicols}

We begin by training the Random Forest model with subset 1 containing all 22 features to establish an initial control output. From there, we re-train the model for each feature removed in order to find the best subset. This is repeated until there is no improvement in the $f_1$ score or until the features are all exhausted. From Figure \ref{backward-elimination}, we can see that there are 7 subsets that have achieved the highest possible $f_1$ score ($f_1\text{ score}=1.0$). Because of this, we stop at 21 features instead of removing more as we cannot achieve a higher $f_1$ score. Therefore, the best subsets are: subset 5, subset 7, subset 11, subset 12, subset 13, subset 17 and subset 22. 

\paragraph{Correlation}~~\\
Feature Correlation is useful to identify how each feature relates to one another. Correlation can either be positive or negative. A positive correlation indicates that an increase in one value of a feature increases the value of the other feature, while the latter indicates the opposite. Both positive and negative correlation can indicate highly correlated features. The closer the score is to the value 1 indicates a high positive correlation, and the closer the score is to the value -1 indicates a high negative correlation. Below is a heatmap to illustrate the correlation among all our features. 

\begin{figure}[H]
\centering
    \includegraphics[width=0.90\textwidth]{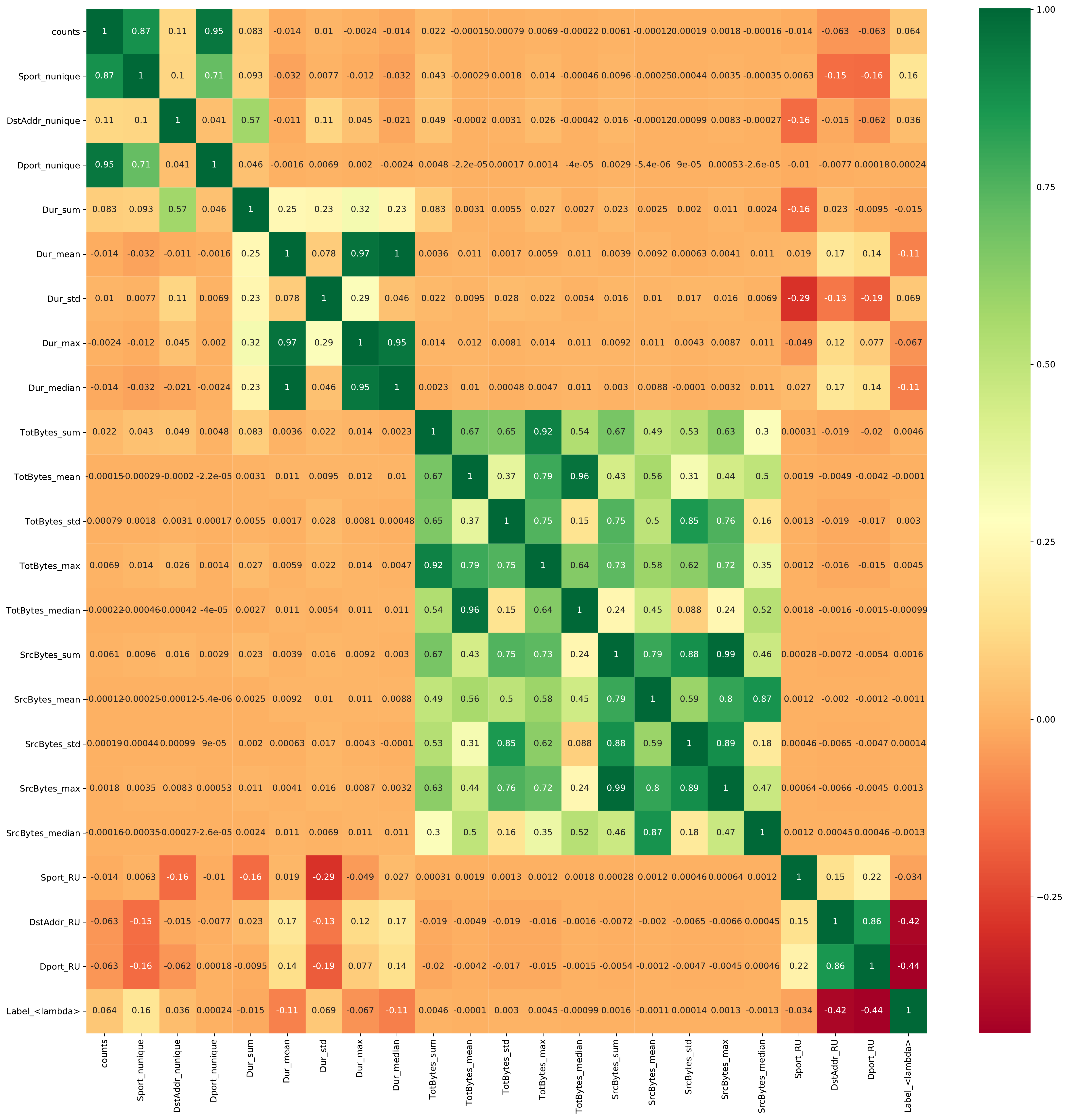}
    \caption[figure]{Feature Correlation Heatmap}
    \label{feature_correlation_heatmap}
\end{figure}

From Figure \ref{feature_correlation_heatmap}, we can deduce which of the features have the highest correlation among each other. The green colored tiles indicate the highest positive correlations, whereas the red ones indicate the highest negative correlations. High correlation is observed in all of the Duration, \texttt{TotBytes}, \texttt{SrcBytes} and entropy features, but this is not surprising as they are derived from statistical measures of the same column in the raw dataset. \\

Relating back to the filter method, we can now use the heatmap to identify highly correlated features within the subset containing the best features and select the best ones that represent it. From the heatmap, it is seen that features \texttt{Dport\_RU} and \texttt{DstAddr\_RU}, as well as features \texttt{Dur\_mean} and \texttt{Dur\_median}, are highly correlated. As a result, we select the features with the higher correlation score with respect to the target label. The end result of the filter method then yields the final subset of best features which contains: \texttt{Dport\_RU}, \texttt{Sport\_nunique} and \texttt{Dur\_median}.

\subsubsection{Embedded methods}
\label{feature_selection2}
Embedded methods can also be used to select features according to the score of the classifier (ie using labels).

\paragraph{Lasso and Ridge logistic regression}~~\\
\label{weight_detection}
In order to use Logistic Regression, we need to find a way to cope with the data imbalance. Thus, cross validation is used to find the best weight to balance the dataset.
\begin{figure}[H]
\centering
    \includegraphics[trim={11 10 12 13}, clip, width=0.73\textwidth]{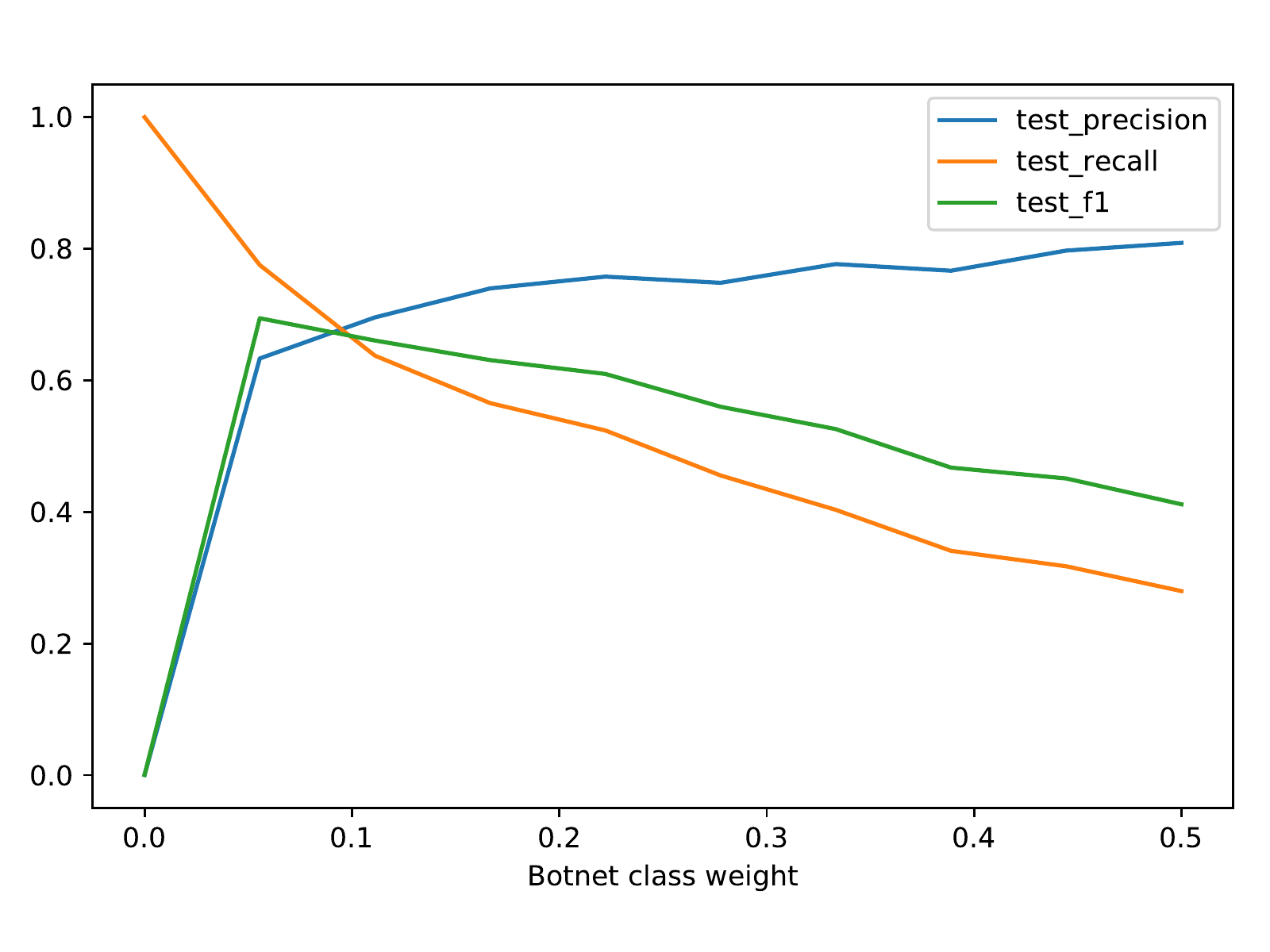}
    \caption[figure]{Scores wrt Botnet class weight}
    \label{cross_validation_class_weight1}
\end{figure}
\begin{figure}[H]
\centering
    \includegraphics[trim={11 20 12 30}, clip, width=0.73\textwidth]{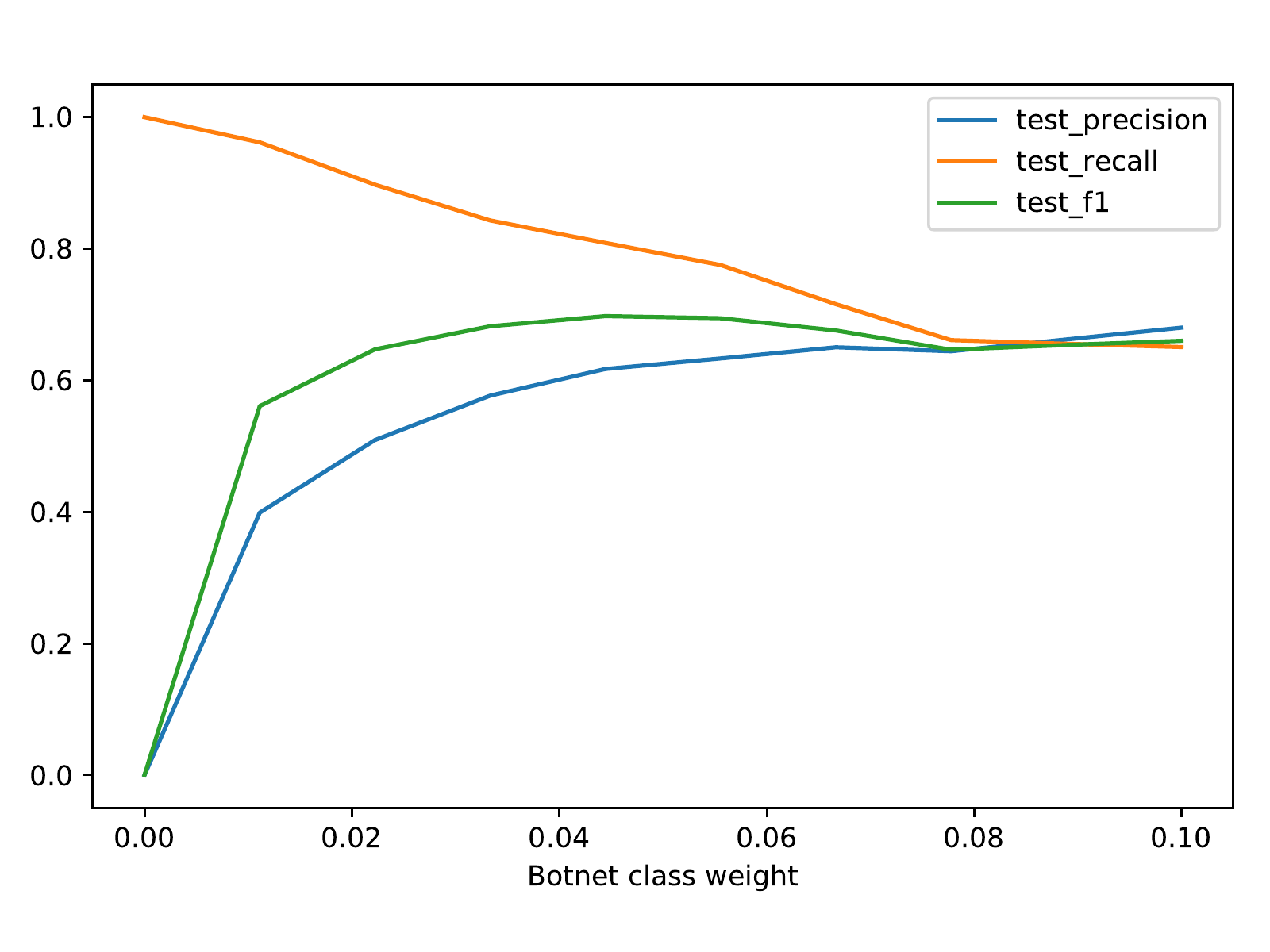}
    \caption[figure]{Scores wrt Botnet class weight (Zoom)}
    \label{cross_validation_class_weight2}
\end{figure}
As shown in Figures \ref{cross_validation_class_weight1} and \ref{cross_validation_class_weight2}, the best weight to have a high $f_1$ score (see Section \ref{section_metrics} to understand the chosen metric) is 0.044. This will be used to balance the amount of botnet data with the amount of background data.\\

Now, a $l_2$ regularization is used to evaluate the most relevant features to maximize the $f_1$ score (Ridge method). In order to do this, the coefficient $C$, inversely proportional to the regularization strength, is tuned.

\begin{figure}[H]
\centering
    \includegraphics[trim={11 20 12 30}, clip, width=0.73\textwidth]{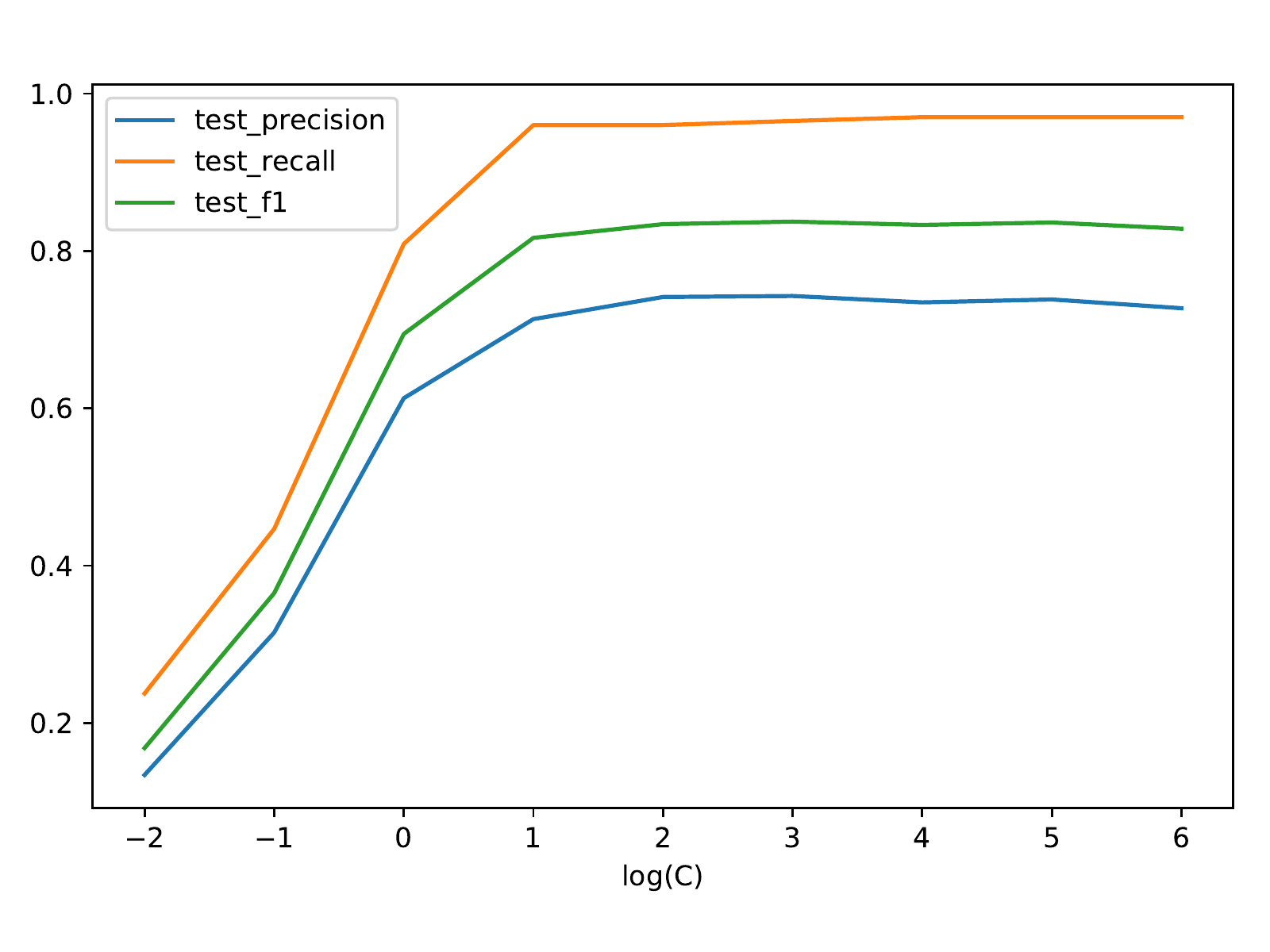}
    \caption[figure]{Scores wrt log(C)}
    \label{cross_validation_C1}
\end{figure}
\begin{figure}[H]
\centering
    \includegraphics[trim={11 10 12 13}, clip, width=0.73\textwidth]{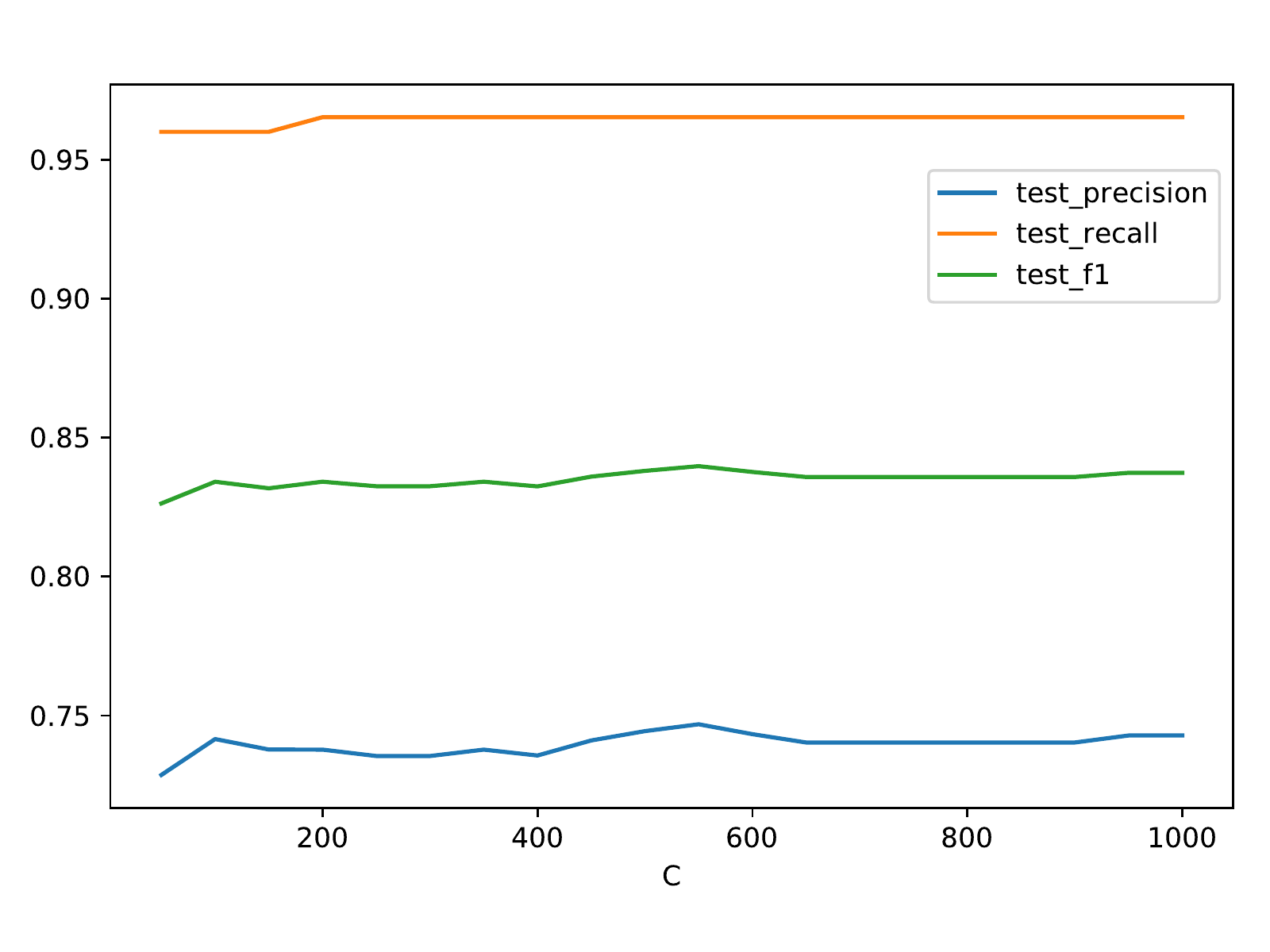}
    \caption[figure]{Scores wrt C (Zoom)}
    \label{cross_validation_C2}
\end{figure}
Figures \ref{cross_validation_C1} and \ref{cross_validation_C2} show the evolution of the different scores with respect to the regularization parameter $C$. The best $f_1$ score is obtained for $C=550$ so the feature coefficients must be analyzed with this parameter.\\

As we are interested in the feature coefficients, here is the detailed list of the features:
\begin{enumerate}
    \setlength{\columnsep}{28pt}
    \begin{multicols}{2}
    \small
    \setcounter{enumi}{-1}
    \item The number of communications in the subgroup
    \item The number of unique source ports
    \item The number of unique destination addresses
    \item The number of unique destination ports
    \item The sum of the communication durations
    \item The mean of the communication durations
    \item The standard deviation of the communication durations
    \item The maximum of the communication durations
    \item The median of the communication durations
    \item The sum of the total number of exchanged bytes
    \item The mean of the total number of exchanged bytes
    \item The standard deviation of the total number of exchanged bytes
    \item The maximum of the total number of exchanged bytes
    \item The median of the total number of exchanged bytes
    \item The sum of the number of bytes sent by the source
    \item The mean of the number of bytes sent by the source
    \item The standard deviation of the number of bytes sent by the source
    \item The maximum of the number of bytes sent by the source
    \item The median of the number of bytes sent by the source
    \item The normalized entropy of the source ports
    \item The normalized entropy of the destination addresses
    \item The normalized entropy of the destination ports
    \end{multicols}
\end{enumerate}

\begin{figure}[H]
\centering
    \includegraphics[trim={11 7 10 11}, clip, width=0.73\textwidth]{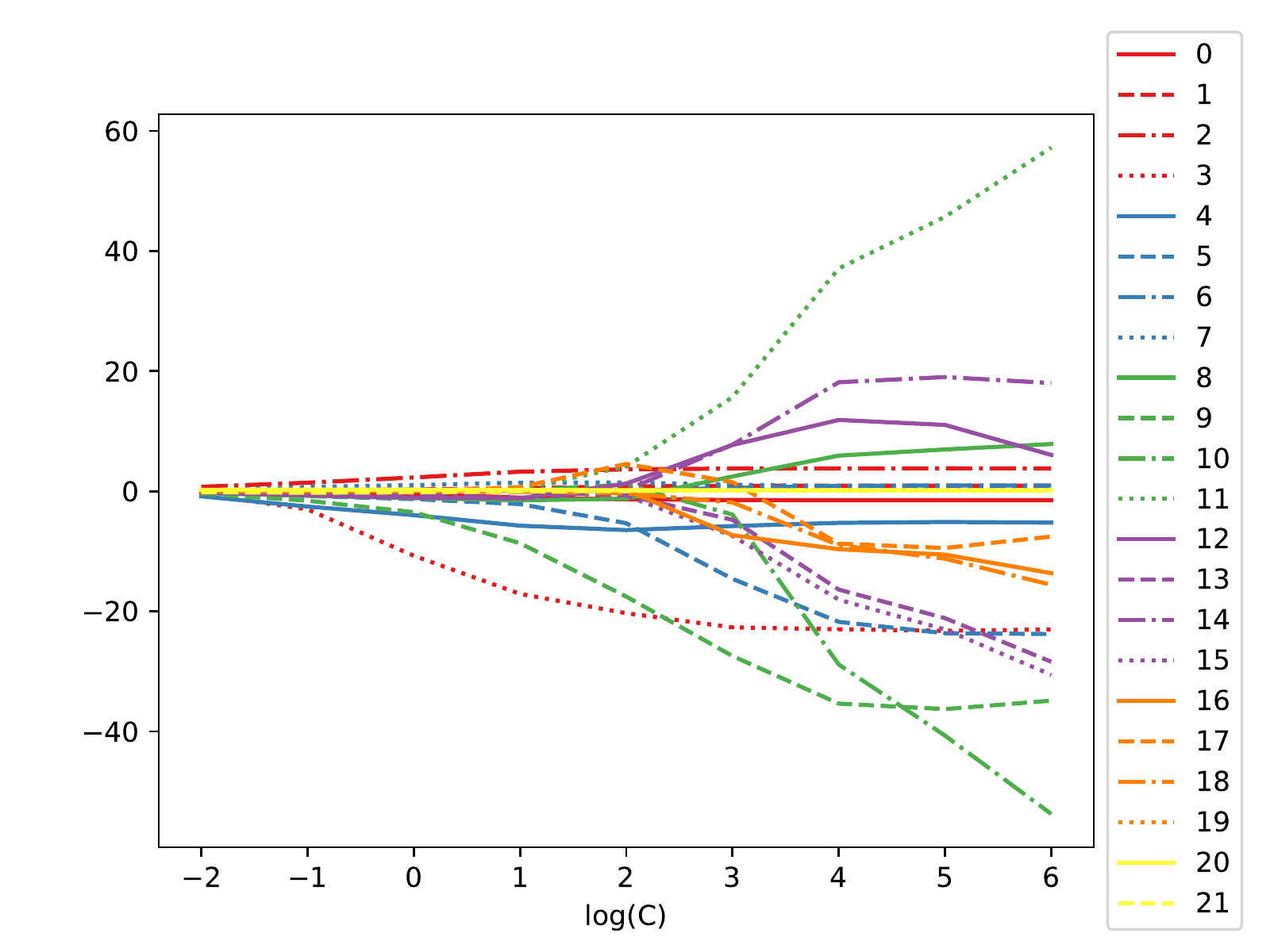}
    \caption[figure]{Feature coefficients wrt log(C)}
    \label{cross_validation_C_coeff1}
\end{figure}
\begin{figure}[H]
\centering
    \includegraphics[trim={11 7 10 11}, clip, width=0.73\textwidth]{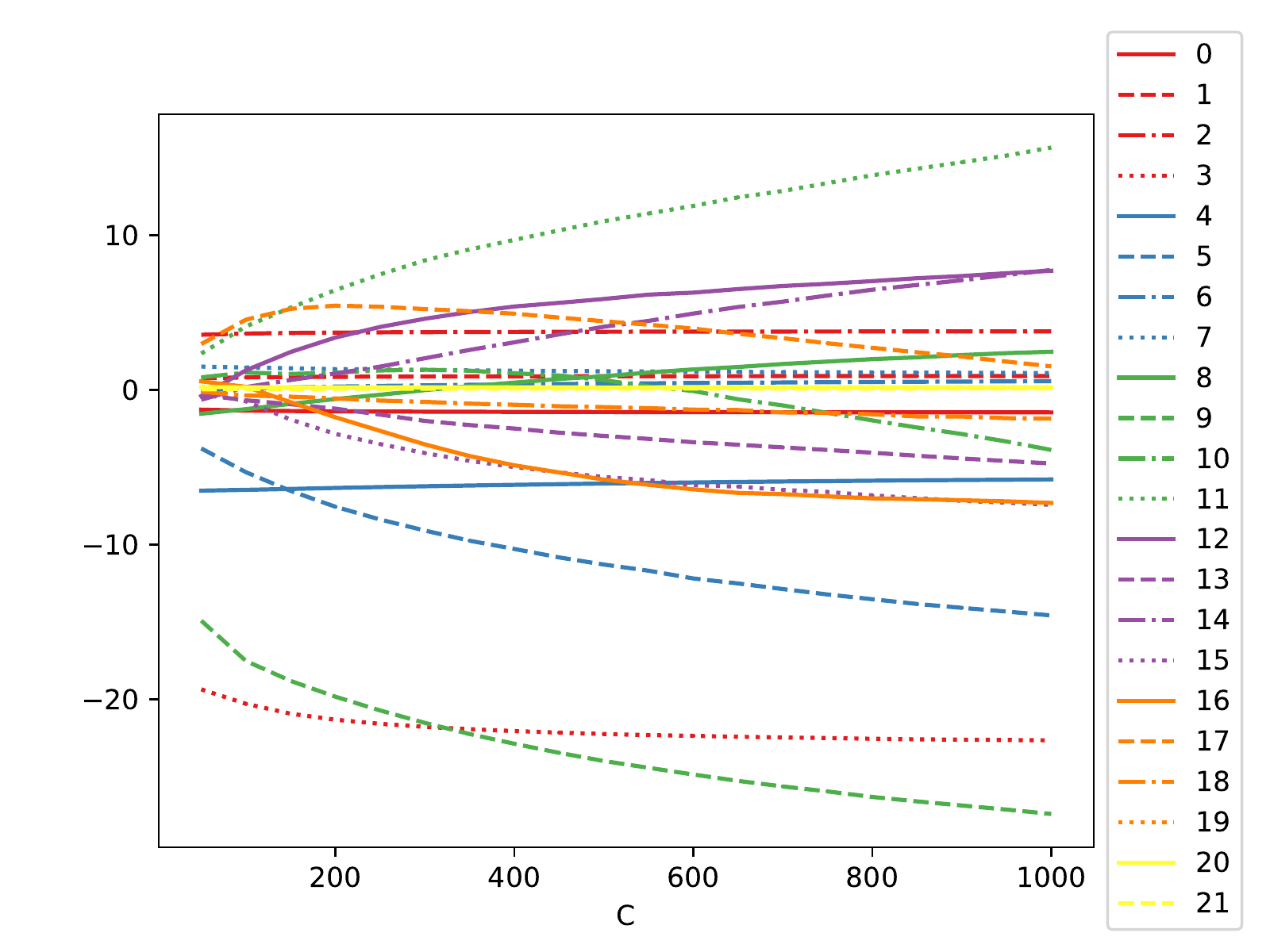}
    \caption[figure]{Feature coefficients wrt C (Zoom)}
    \label{cross_validation_C_coeff2}
\end{figure}
Figures \ref{cross_validation_C_coeff1} and \ref{cross_validation_C_coeff2} represent the importance of each feature with respect to $C$. We can see that the features 3 and 9 have the most impact on the results. Then, the features 4, 5 and 11 seem to contribute to the score as well as the features 12 and 16 for $C=550$. On the other hand, the features 1, 6, 10, 19, 20, 21 do not have a strong impact on the results and can easily be forgotten.\\

Now, the same thing is done with a $l_1$ regularization (Lasso method). This technique is more computational expensive because there is no analytic solution, so the score tolerance has to be reduced. Nevertheless, using the solver SAGA (Stochastic Average Gradient Augmented), the algorithm did not succeed to converge, giving thus very poor results. Consequently, despite being slower, the solver Liblinear (A Library for Large Linear Classification) is used. However, this technique has not succeeded to give relevant results since it cannot perform cross-validation over the parameter $C$ (very slow to converge).

\paragraph{Support Vector Machine (SVM) method with Recursive Feature Elimination (RFE)}~~\\
Another technique to select features is to recursively remove the less relevant feature and train the model to analyze the scores. For this, we used the Support Vector Machine Classification algorithm with the Stochastic Gradient Descent (SGD) learning method.\\

\begin{figure}[H]
\centering
    \includegraphics[trim={11 7 10 11}, clip, width=0.73\textwidth]{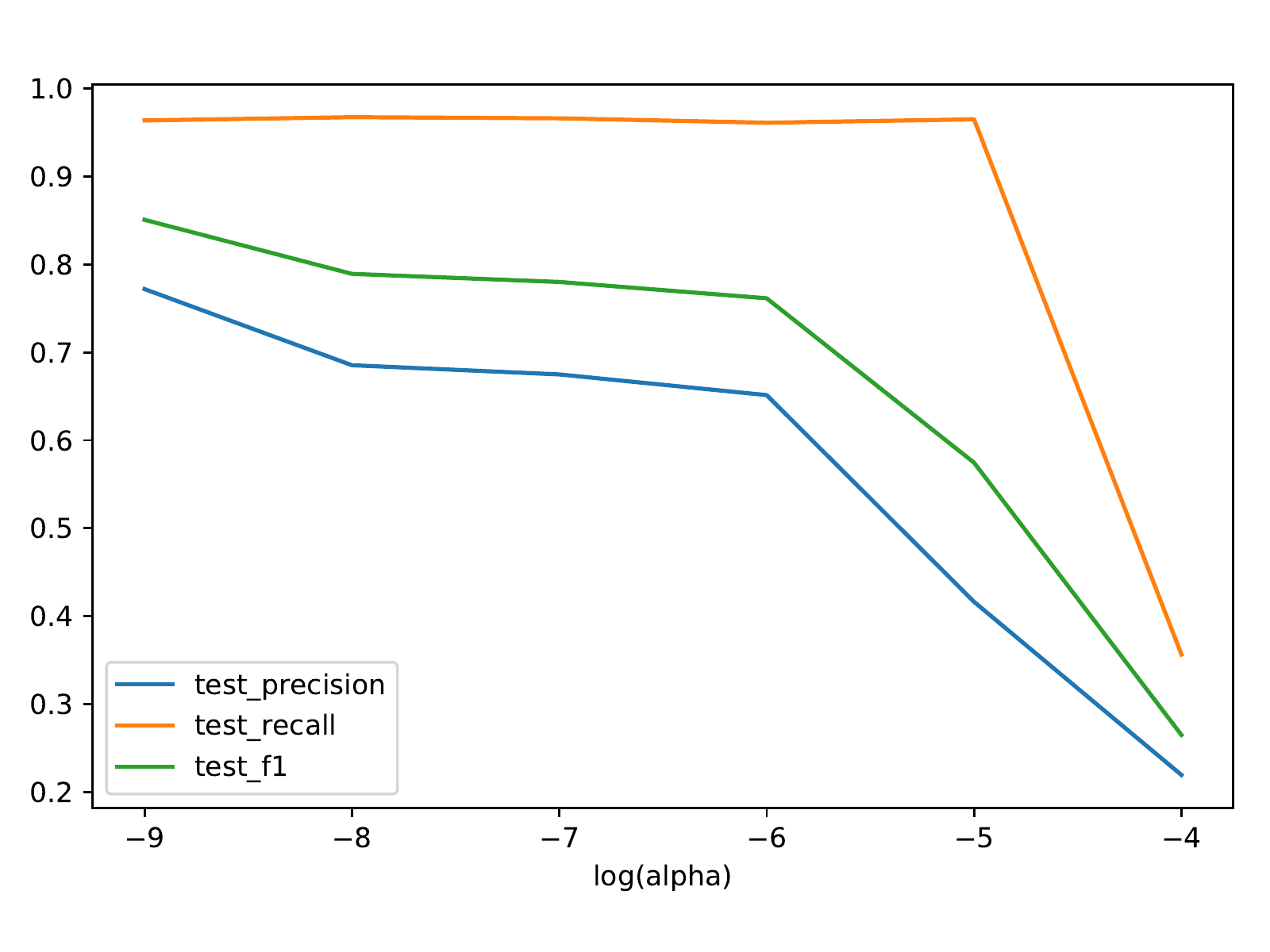}
    \caption[figure]{Scores wrt alpha (Part 1)}
    \label{cross_validation_alpha1}
\end{figure}

\begin{figure}[H]
\centering
    \includegraphics[trim={11 7 10 11}, clip, width=0.73\textwidth]{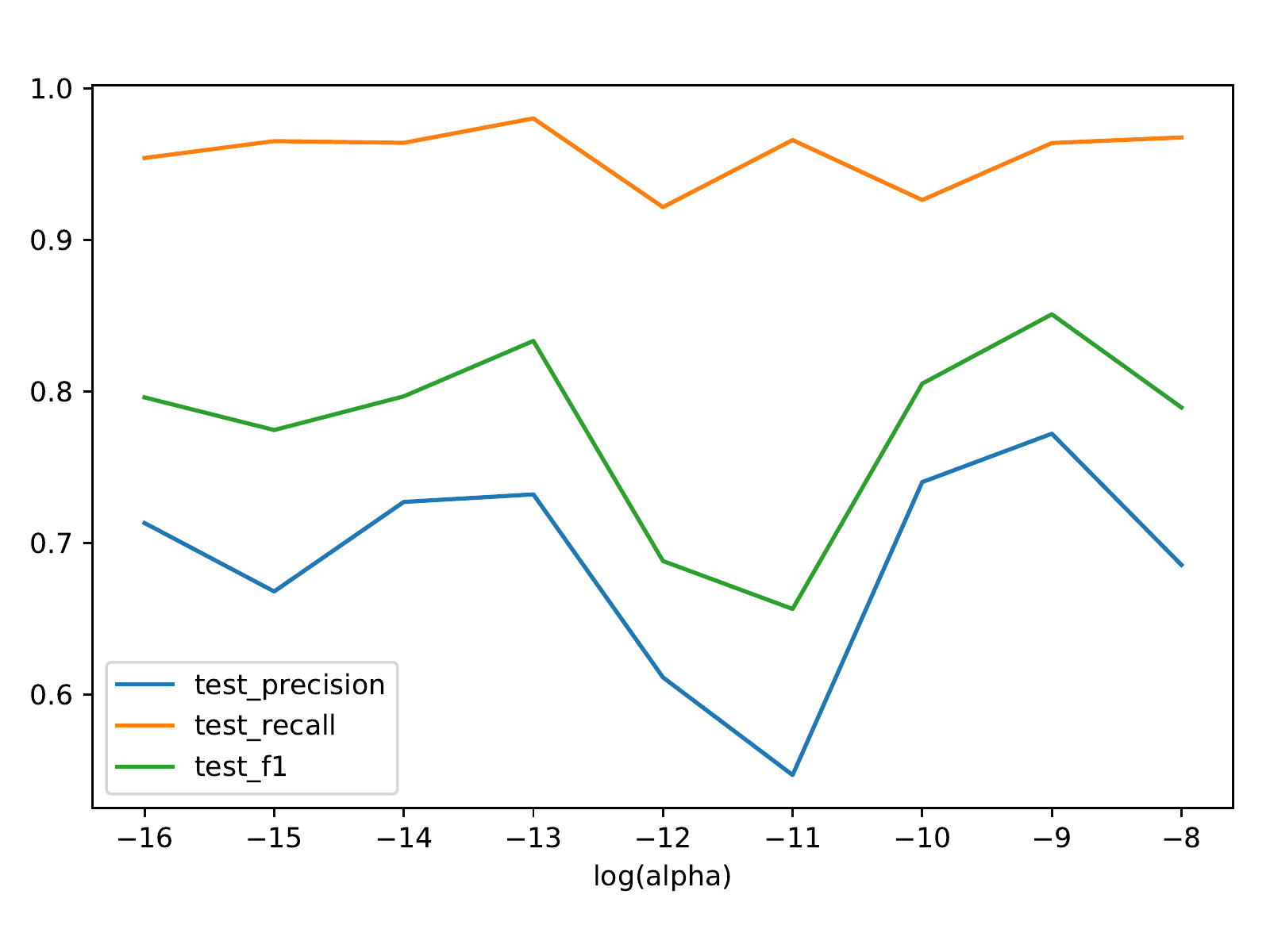}
    \caption[figure]{Scores wrt alpha (Part 2)}
    \label{cross_validation_alpha2}
\end{figure}

We also used an ElasticNet regularization (using $l_1$ and $l_2$) with a $l_1$ ratio of 15\%, and in order to choose the regularization parameter $\alpha$, we used a cross validation method. Figures \ref{cross_validation_alpha1} and \ref{cross_validation_alpha2} show that the best value for $\alpha$ is $10^{-9}$. They also suggest that a too low regularization parameter leads to a lot of variability in the results.

The results from the Recursive Feature Elimination were not relevant enough to enable extracting features because the process of training the model from scratch with less features was time-consuming.

\begin{figure}[H]
\centering
    \includegraphics[width=0.85\textwidth]{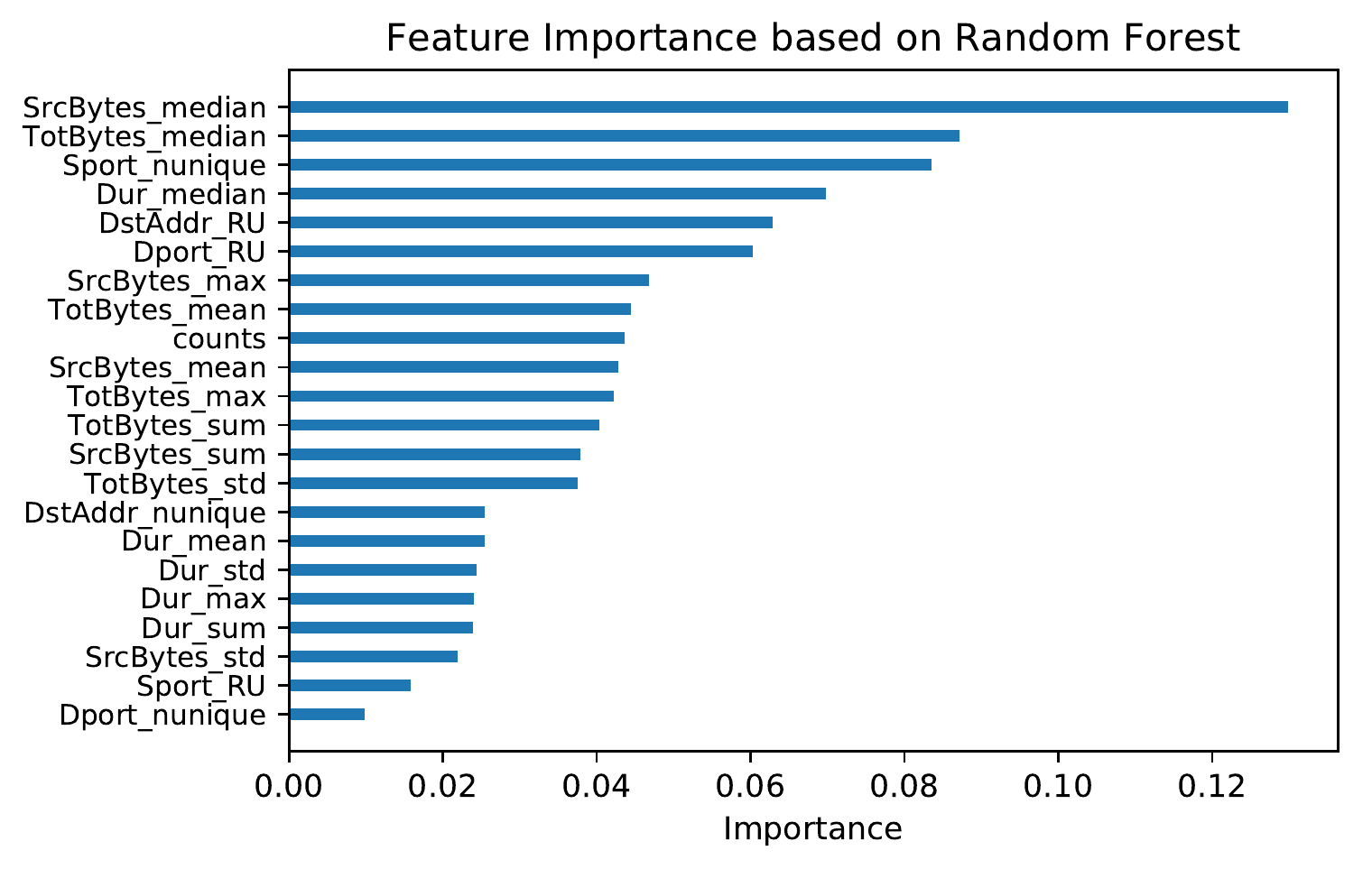}
    \caption[figure]{Feature Importance from Random Forest}
    \label{rf_feature_importance}
\end{figure}

\newpage
\vfill
\paragraph{Feature Importance of Random Forest Classifier}~~\\
Random Forest Classifier has a built-in attribute (\texttt{feature\_importances\_}) that returns an array of values, with each value representing the measure of importance of a feature. The value is determined based on how each feature contributes to decreasing the weighted Gini impurity \footnote{Gini impurity is a measurement used in Classification and Regression Trees (CART) to measure the likelihood of an incorrect classification.}. The higher the importance value, the more significant that feature is. Figure \ref{rf_feature_importance} illustrates the importance for all of the extracted features.

From Figure \ref{rf_feature_importance}, we can deduce that the most important features based on the Random Forest algorithm are \texttt{SrcBytes\_median}, \texttt{TotBytes\_median}, \texttt{Sport\_nunique} and \texttt{Dur\_median}.
\vfill
To summarize, there are different techniques of feature selection but, as you can see, they contradict each other (The number of unique destination ports for example is a very important feature with $l_2$ Logistic Regression, but a very insignificant with the Random Forest Classifier attribute). As a consequence, the final models are trained with all the extracted features (22 features).

\newpage
\subsubsection{Dimensionality reduction}
Another idea to reduce the number of features is to use dimensionality reduction techniques. It is also very useful to visualize the dataset.

\paragraph{Principal Component Analysis (PCA)}~~\\
The Principal Component Analysis method enables the reduction of the dataset dimension and returns a set of linearly uncorrected variables.
\begin{figure}[H]
\centering
    \includegraphics[trim={11 7 10 11}, clip, width=0.73\textwidth]{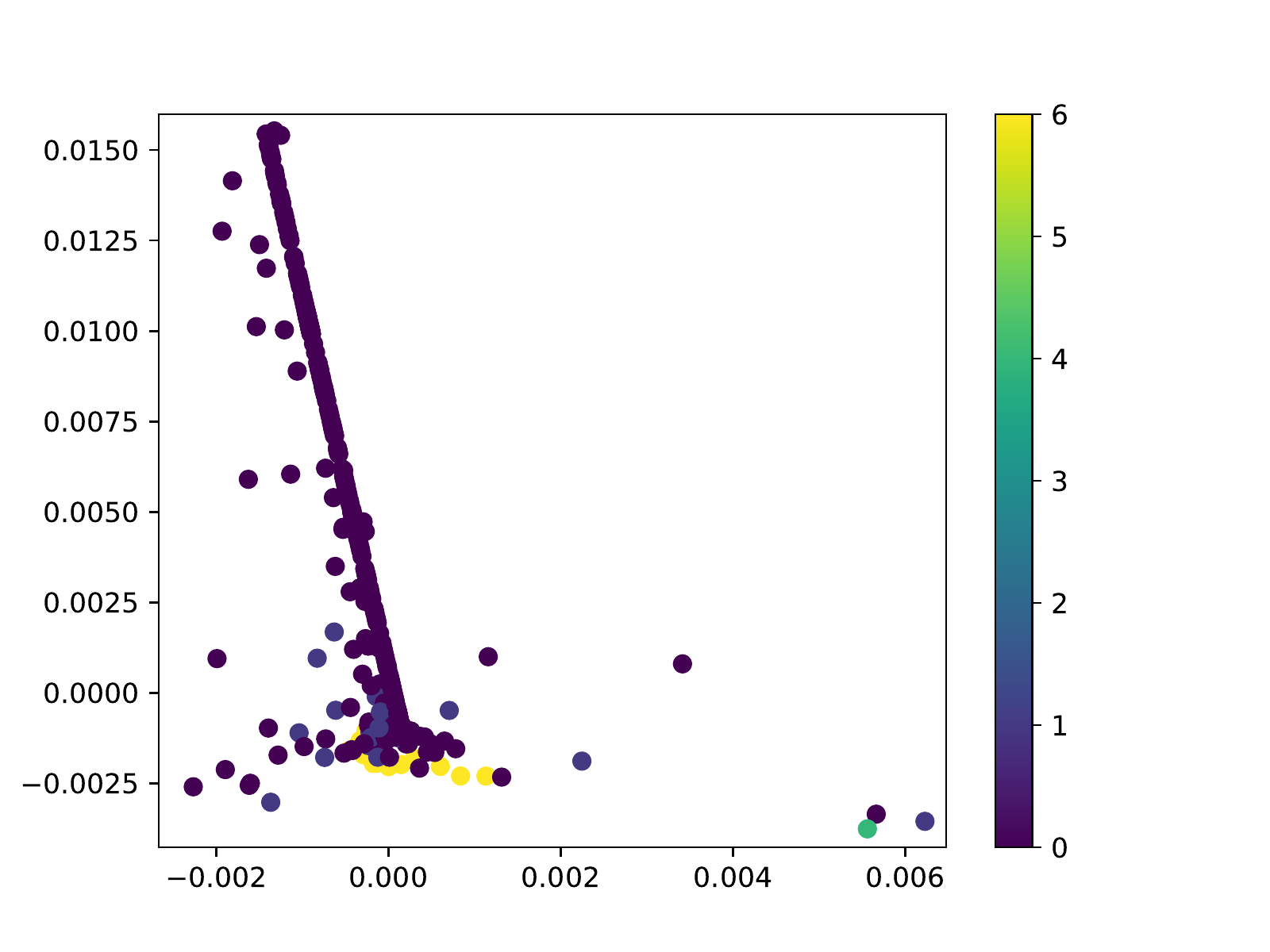}
    \caption[figure]{Principal Component Analysis}
    \label{pca}
\end{figure}
Figure \ref{pca} shows the representation of the first two components of PCA, with the set of pre-selected features \{3, 4, 5, 9, 11\}. The botnets are labeled as 6 and some non-botnet points are not displayed\footnote{The number of displayed non-botnet points is a hundred times larger than the number of botnet points.}. The first component (vertical axis) represents 58\% of the dataset variance and the second component (horizontal axis) explains 35\% of the total variance.\\

We can see that the botnet points get into a group at the bottom of a long strait line of background points. This representation can be used to train a K-Nearest Neighbours classifier.

\newpage
\paragraph{t-SNE (t-distributed Stochastic Neighbour Embedding)}~~\\
T-SNE is another method to reduce the dimension of the dataset to 2 or 3 features.
\begin{figure}[H]
\centering
    \includegraphics[trim={88 35 165 48}, clip, width=1\textwidth]{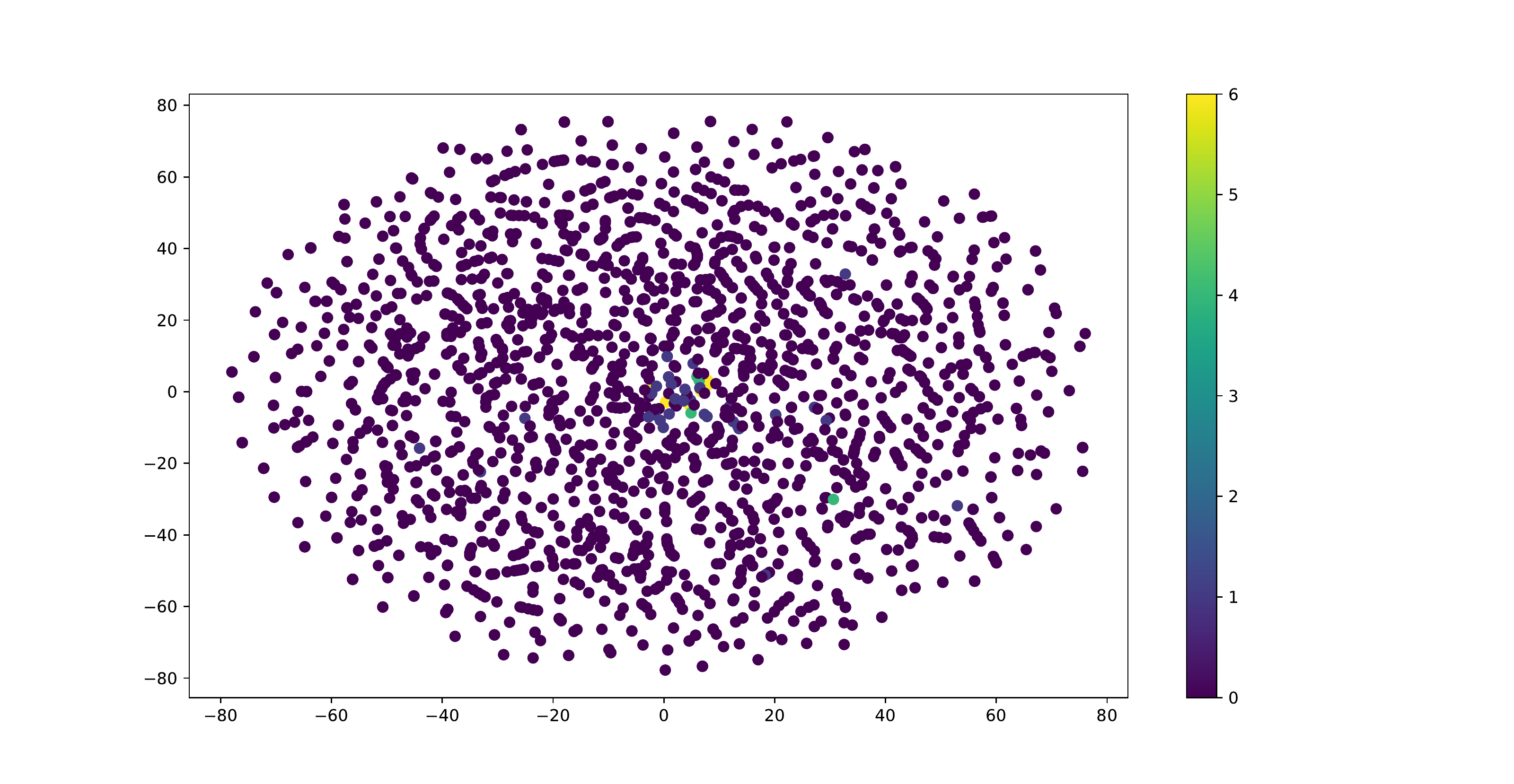}
    \caption[figure]{t-SNE}
    \label{tsne}
\end{figure}
Figure \ref{tsne} shows the result of the dimensionality reduction (the botnets are labeled as 6). We can see that the botnet points are not gathered in one place that could differentiate them from the regular traffic. Moreover, the algorithm is very time and memory consuming. Consequently, the use of the K-nearest neighbours method is no longer taken into consideration to detect botnet activities in this paper.

\section{Results}
\subsection{Metrics}
\label{section_metrics}
In order to compare the results of different algorithms, some metrics need to be chosen to evaluate the performance of the methods. The usual way is to look at the number of false positives (ie the number of background communications labeled as botnets) and false negatives (ie the number of botnets labeled as background communications).\\

To do this, three scores can be used\footnote{tp: true positives; fp: false positives; fn: false negatives}:
\begin{itemize}
    \item the Recall: $\displaystyle R = \frac{tp}{tp + fn}$
    \item the Precision: $\displaystyle P = \frac{tp}{tp + fp}$
    \item the $f_1$ Score: $\displaystyle f_1 = \left(\frac{R^{-1}+P^{-1}}{2}\right)^{-1} = 2 \times \frac{R \times P}{R+P}$
\end{itemize}

One needs to remember that, for our project of detecting malicious software, having a low recall is worst than having a low precision because it means that most detected communications are botnets (precision) but most botnet communications remain undetected (recall).

Of course, having a good recall is easy because all you have to do is label every communication as botnet. So the chosen compromise is to maximize the $f_1$ score while checking that the recall is not too low.\\

\underline{Remark:} The accuracy score, defined as the normalized number of well predicted labels, is not relevant to our case because of the imbalance of the dataset. Only a weighted accuracy can be relevant (using the weight found in Section \ref{weight_detection} with the $f_1$ score).

\subsection{Algorithms}
\subsubsection{Logistic Regression}
The Logistic Regression method is an algorithm that uses a linear combination of the features to classify the flows. As presented in Section \ref{weight_detection}, cross-validation is necessary to tune the parameters of the model. The final chosen parameters are $C=550$ and $\text{Weight}_{\text{non-botnet}}=0.044$.

\subsubsection{Support Vector Machine}
The Support Vector Machine method is an algorithm that uses kernels to transform the data space and then try to find a separate line to split the data into two classes. As presented in Section \ref{weight_detection}, cross-validation is necessary to tune the parameters of the model. For a linear kernel, the chosen $\alpha$ parameter is $\alpha = 10^{-9}$.

\begin{figure}[H]
\centering
    \includegraphics[trim={11 20 10 30}, clip, width=0.73\textwidth]{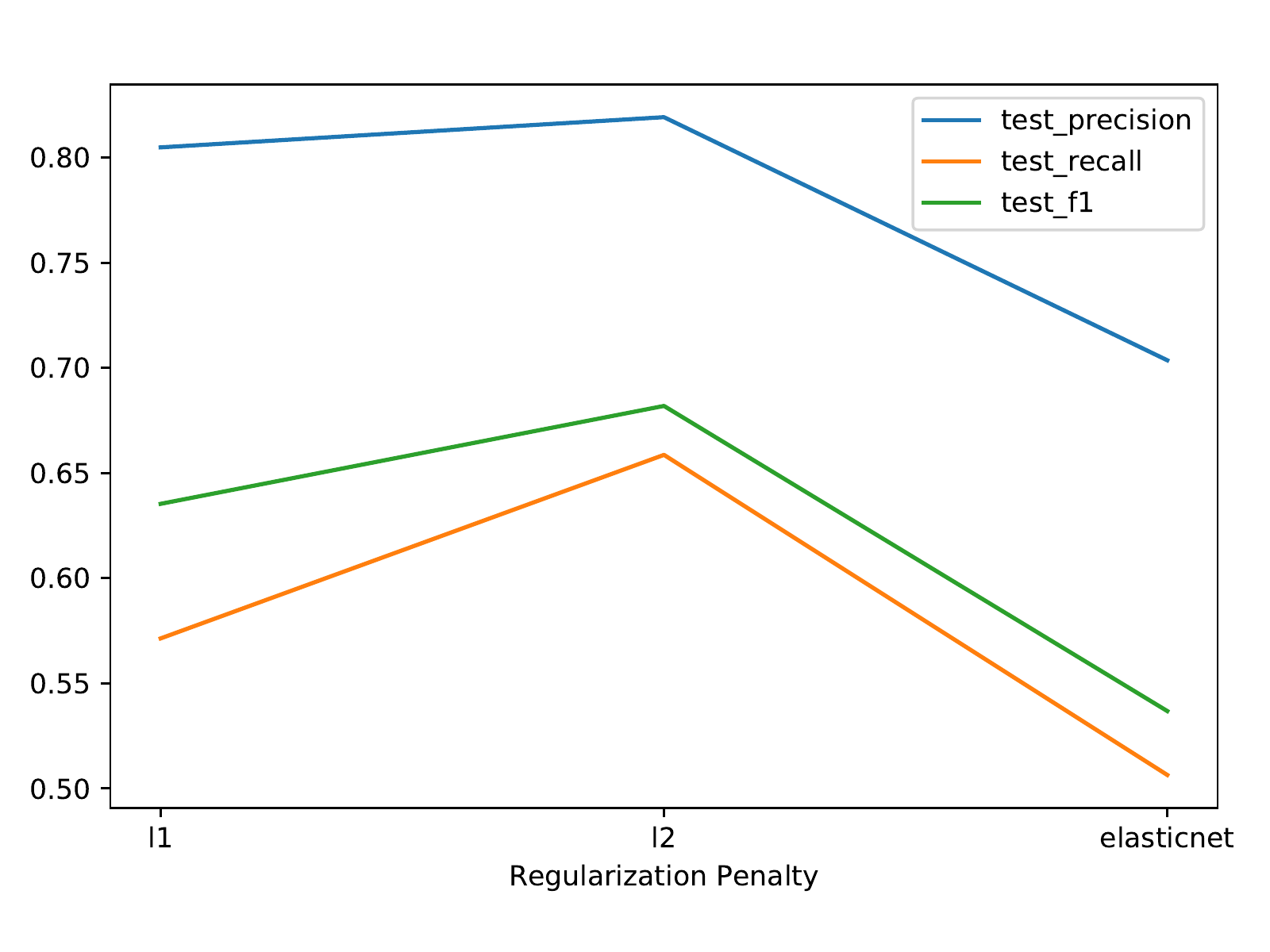}
    \caption[figure]{Scores wrt Regularization Penalty}
    \label{svm_penalty}
\end{figure}
In Figure \ref{svm_penalty}, cross-validation is performed to find the best regularization method. The results show that a $l_2$ regularization is preferred.

\begin{figure}[H]
\centering
    \includegraphics[trim={11 20 10 30}, clip, width=0.73\textwidth]{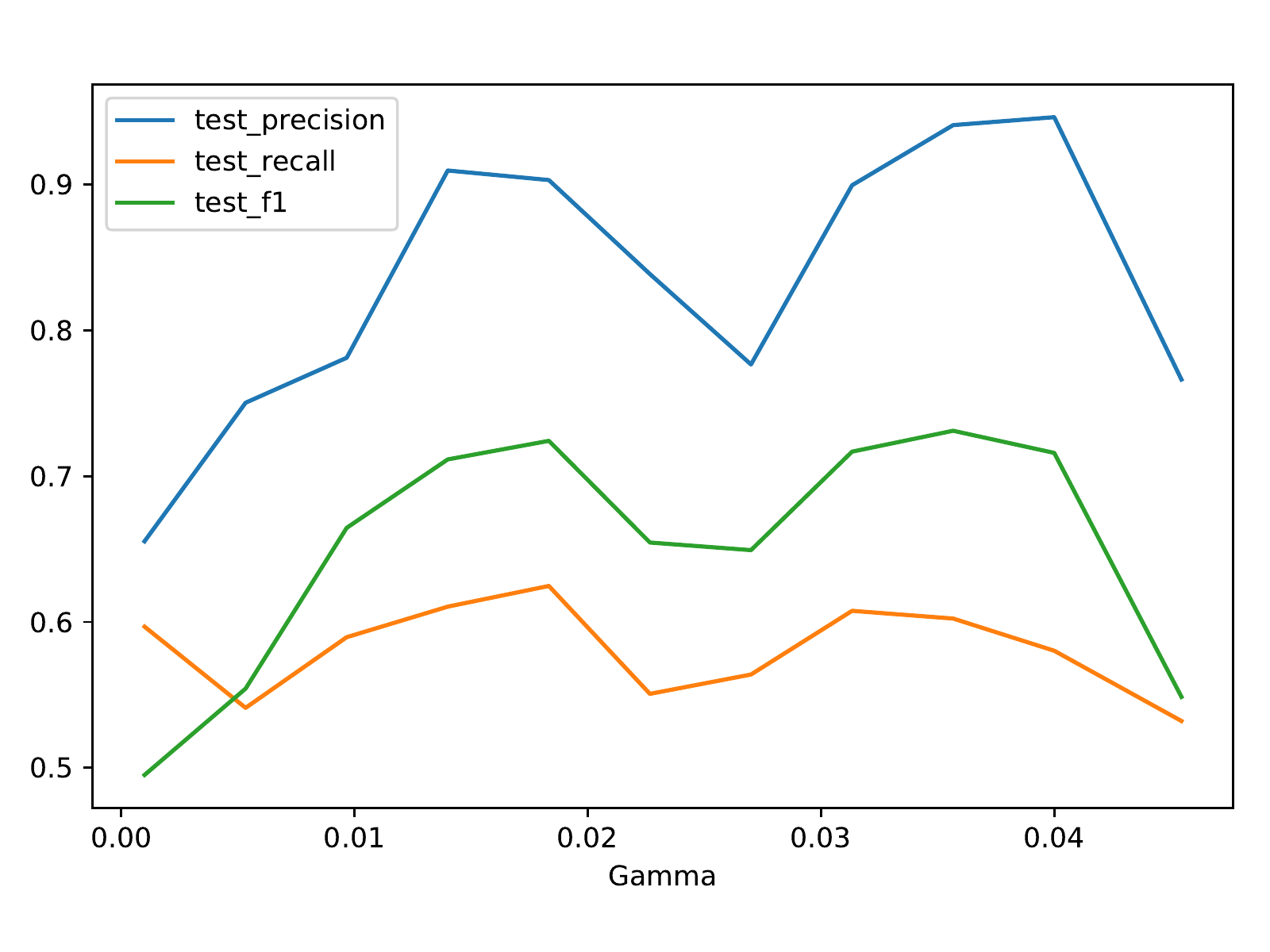}
    \caption[figure]{Scores wrt Gamma}
    \label{svm_gamma}
\end{figure}
For a Radial Basis Function (RBF) kernel, Figure \ref{svm_gamma} shows that the best $\gamma$ parameter is $\gamma=0.03567$.

\begin{figure}[H]
\centering
    \includegraphics[trim={11 20 10 30}, clip, width=0.73\textwidth]{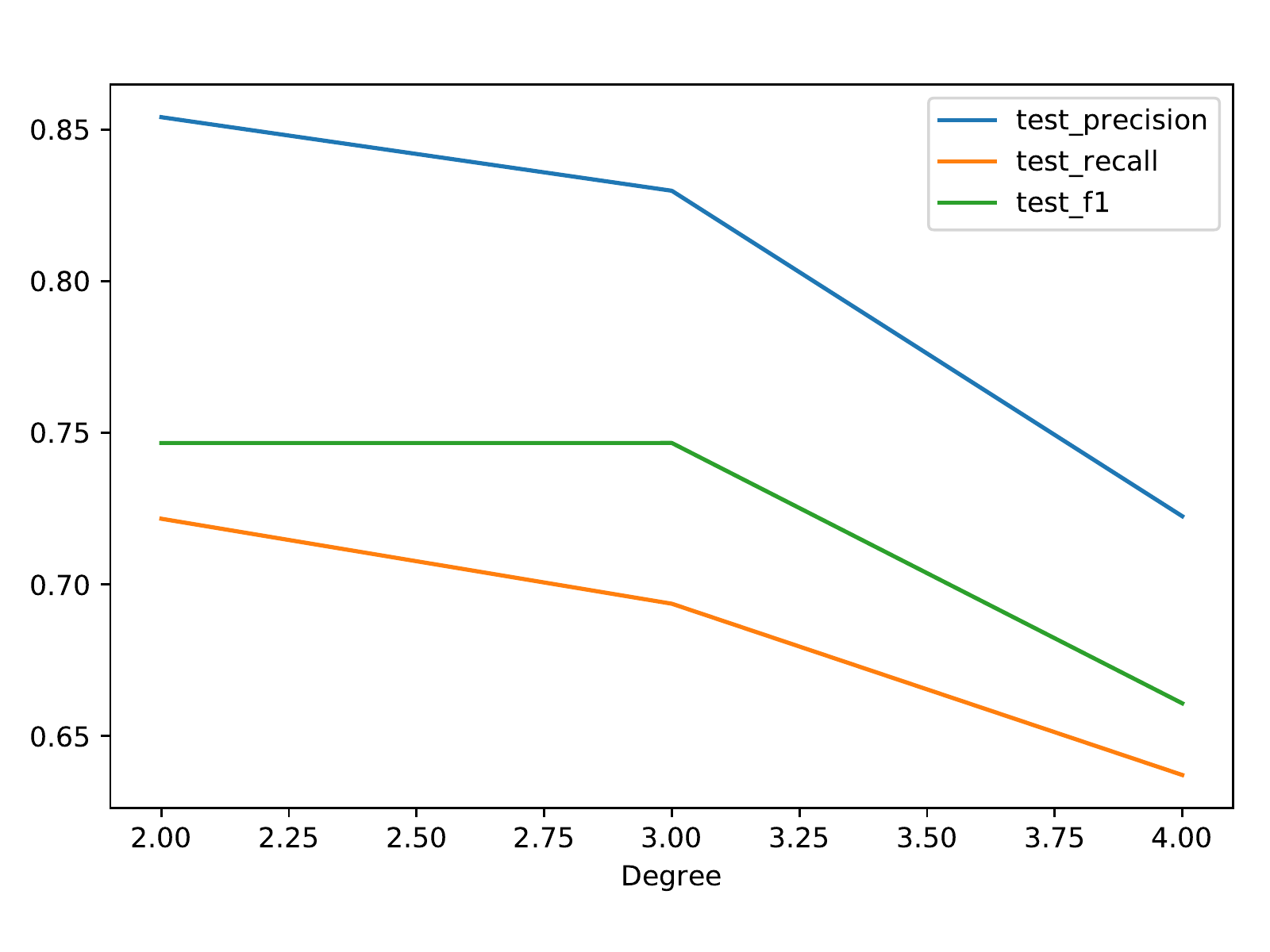}
    \caption[figure]{Scores wrt Polynomial Degree}
    \label{svm_degree}
\end{figure}
Finally, for a polynomial kernel, Figure \ref{svm_degree} shows that the best degree for the polynomial function is 2.\\

To conclude, the best SVM parameters for the botnet detection in the first scenario of CTU-13 are a polynomial kernel with a degree of 2.

\subsubsection{Random Forest}
Random Forest is an algorithm that uses several decision tree classifiers to predict the class of each input flows. The chosen number of trees in the forest is 100.

\subsubsection{Gradient Boosting}
The Gradient Boosting method is an algorithm that also uses decision tree classifiers but tries to incrementally improve the training by using the score of one tree to build another one. Two main parameters have to be tuned: the function loss and the maximum depth of the trees (the chosen number of trees is 100.).

\begin{figure}[H]
\centering
    \includegraphics[trim={11 23 10 30}, clip, width=0.67\textwidth]{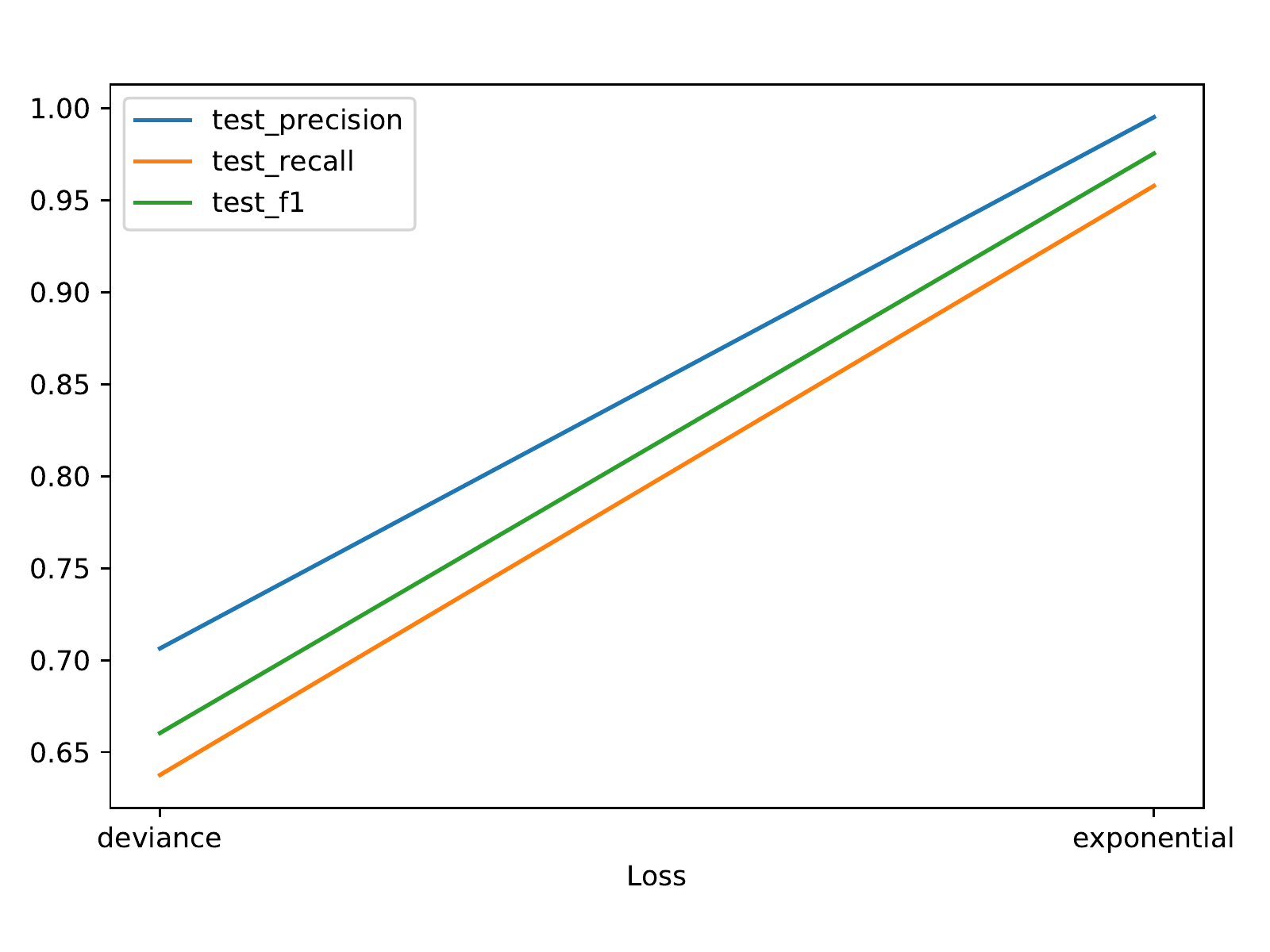}
    \caption[figure]{Scores wrt chosen Loss}
    \label{gboost_loss}
\end{figure}
Figure \ref{gboost_loss} shows that an exponential loss performs better than a deviance loss.

\begin{figure}[H]
\centering
    \includegraphics[trim={11 23 10 30}, clip, width=0.67\textwidth]{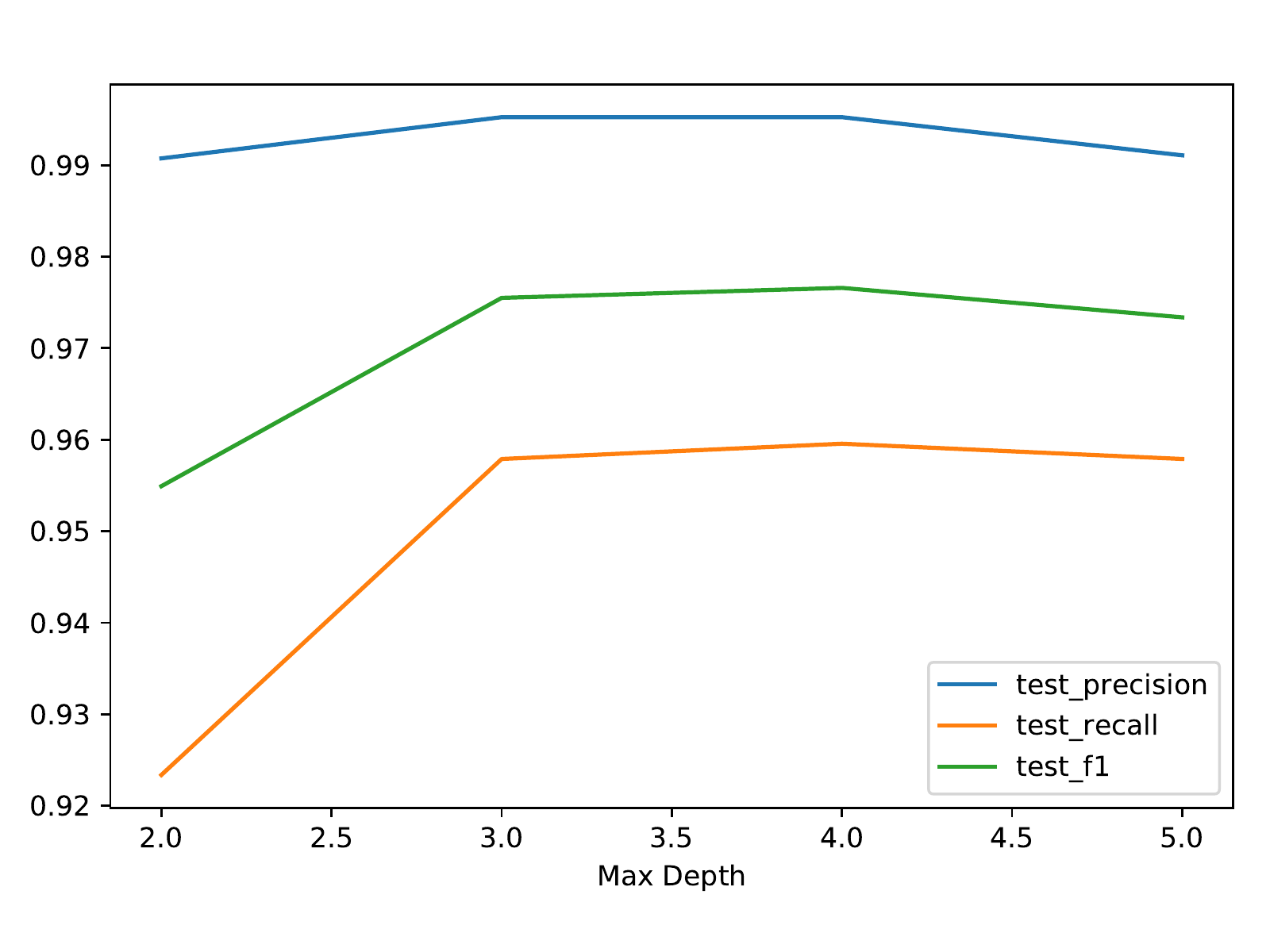}
    \caption[figure]{Scores wrt Max depth}
    \label{gboost_max_depth}
\end{figure}
Figure \ref{gboost_max_depth} shows that the most relevant maximum depth for the classifier is 4.

\subsubsection{Dense Neural Network}
Recently, neural networks have gained popularity since they perform very well when a lot of data is available. We test here a simple dense (or fully connected) neural network with 2 hidden layers: the first one has 256 neurons and the second one has 128.\\

The parameters of the neural network are composed of a batch-normalization, no dropout, a ReLU activation function (except for the output layer where a sigmoid function is used). The model has 39 681 trainable parameters and 768 non-trainable parameters.\\

\begin{figure}[H]
\centering
    \includegraphics[trim={11 7 10 11}, clip, width=0.73\textwidth]{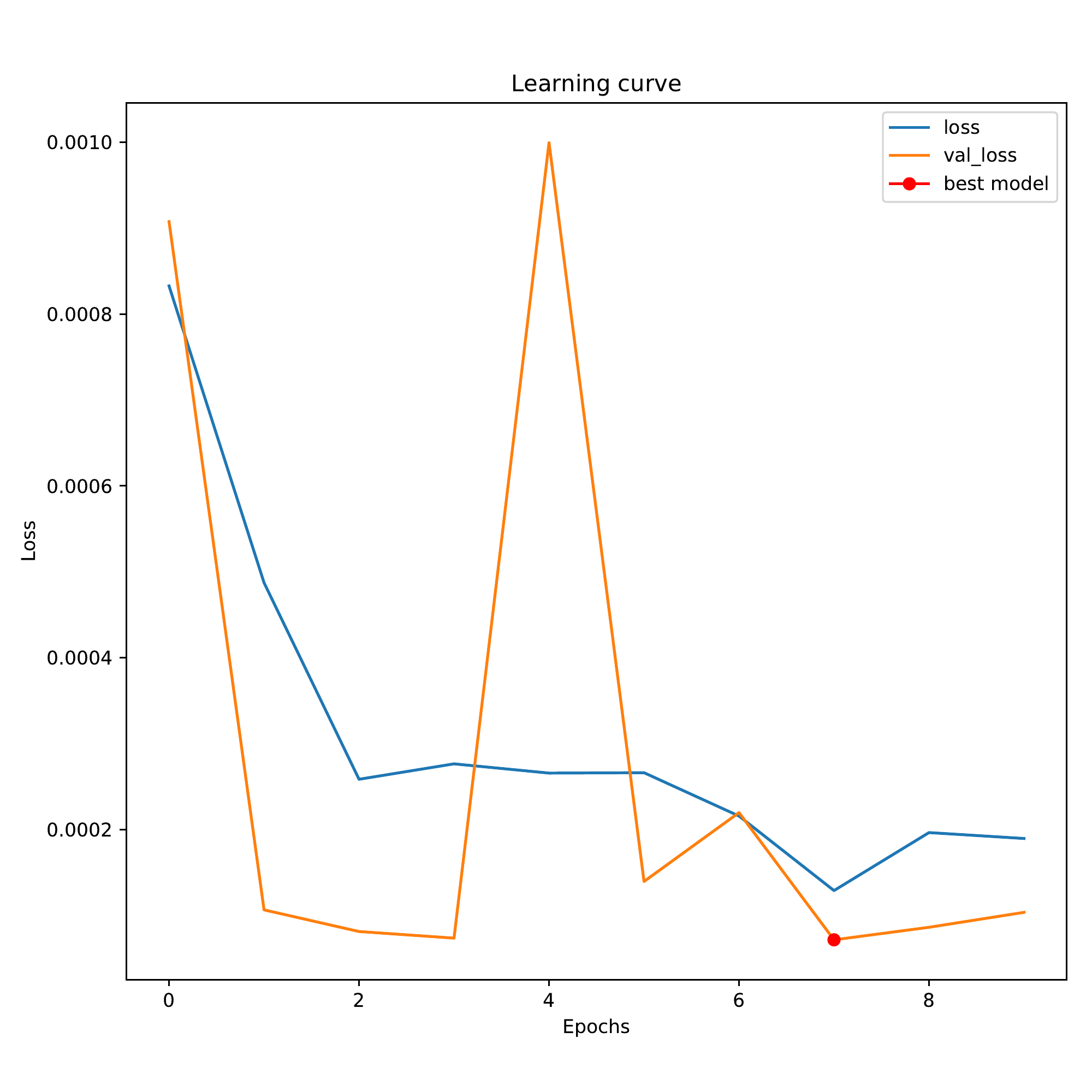}
    \caption[figure]{Learning curve of the Neural Network}
    \label{learning_curve}
\end{figure}

Figure \ref{learning_curve} shows the binary cross-entropy loss for the training set and the validation set (15\% of the whole set) along the 10 epochs (a batch of 32 is used).

\newpage
\subsection{Comparison of algorithms}
In order to compare the algorithms, we train our models with 2/3 of the dataset (NetFlows are randomly chosen) and we test them with the remaining 1/3.
\begin{table}[H]
    \centering
    \caption[table]{Botnet Neris Scenario 1 - Result Summary}
    \label{result_summary_botnet1}
    \setlength\extrarowheight{5pt}
    \begin{tabular}{|c|c|c|c|c|c|c|}
        \hline \multirow{2}{*}{Algorithms} & \multicolumn{3}{c|}{Training} & \multicolumn{3}{c|}{Test} \\
        \cline{2-7}
         & Precision & Recall & $f_1$ Score & Precision & Recall & $f_1$ Score \\
        \hline
        Logistic Regression & 0.74 & 0.97 & 0.83 & 0.74 & 0.96 & 0.84 \\
        \hline
        Support Vector Machine & 0.94 & 0.80 & 0.86 & 0.91 & 0.78 & 0.86 \\
        \hline
        Random Forest & 1.0 & 1.0 & 1.0 & 1.0 & 0.95 & \textcolor{OliveGreen}{0.98} \\
         \hline
        Gradient Boosting & 1.0 & 0.99 & 1.0 & 0.98 & 0.96 & \textcolor{OliveGreen}{0.97} \\
        \hline
        Dense Neural Network & 0.90 & 0.98 & 0.94 & 0.96 & 0.99 & \textcolor{OliveGreen}{0.98} \\
        \hline
    \end{tabular}
\end{table}

Table \ref{result_summary_botnet1} shows that Random Forest, Gradient Boosting and the Dense Neural Network outperform other tested algorithms in detecting botnets among the network traffic with a $f_1$ score of $0.97$. Nevertheless, the Logistic Regression and the Support Vector Machine methods succeed in performing a $f_1$ score of $0.85$, one with a low precision, the other with a low recall.\\

As the Random Forest is the fastest algorithm to train and the less inclined to overfit (because there is not a lot of hyperparameters to choose), this technique will be studied with other scenarios in the next Section.

\newpage
\subsection{Detection of different Botnets}
\subsubsection{Main Results}
In order to see if the results of the Random Forest Classifier on the Scenario 1 of the CTU-13 dataset can be generalized, we test it on all the other scenarios. 

\begin{table}[H]
    \centering
    \caption[table]{Result Summary using Random Forest Classifier}
    \label{result_summary_rfc}
    \setlength\extrarowheight{5pt}
    \begin{tabular}{|c|c|c|c|c|c|c|c|c|}
        \hline \multirow{2}{*}{Botnet} &
        \multicolumn{2}{c|}{Dataset} & \multicolumn{3}{c|}{Training} & \multicolumn{3}{c|}{Test} \\
        \cline{2-9}
         & Size & Botnets & Precision & Recall & $f_1$ Score & Precision & Recall & $f_1$ Score \\
        \hline
        Neris - Scenario 1 & 2 226 720 & 1.28 \textpertenthousand & 1.0 & 1.0 & 1.0 & 1.0 & 0.95 & \textcolor{OliveGreen}{0.975} \\
        \hline
        Neris - Scenario 2 & 1 431 539 & 1.45 \textpertenthousand & 1.0 & 1.0 & 1.0 & 1.0 & 0.98 & \textcolor{OliveGreen}{0.99} \\
        \hline
        Rbot - Scenario 3 & 2 024 053 & 4.99 \textpertenthousand & 1.0 & 0.99 & 0.99 & 1.0 & 0.96 & \textcolor{OliveGreen}{0.98} \\
        \hline
        Rbot - Scenario 4 & 470 663 & 2.36 \textpertenthousand & 1.0 & 0.90 & 0.95 & 1.0 & \textcolor{orange}{0.69} & \textcolor{orange}{0.82} \\
        \hline
        Virut - Scenario 5 & \textcolor{red}{63 643} & 3.46 \textpertenthousand & 1.0 & 1.0 & 1.0 & 1.0 & \textcolor{red}{0.25} & \textcolor{red}{0.4} \\
         \hline
        DonBot - Scenario 6 & 220 863 & 5.57 \textpertenthousand & 1.0 & 1.0 & 1.0 & 1.0 & 0.9 & \textcolor{OliveGreen}{0.95} \\
        \hline
        Sogou - Scenario 7 & \textcolor{red}{50 629} & 1.38 \textpertenthousand & 1.0 & 1.0 & 1.0 & 1.0 & \textcolor{red}{0.25} & \textcolor{red}{0.4} \\
        \hline
        Murlo - Scenario 8 & 1 643 574 & 6.80 \textpertenthousand & 1.0 & 1.0 & 1.0 & 1.0 & 0.94 & \textcolor{OliveGreen}{0.97} \\
        \hline
        Neris - Scenario 9 & 1 168 424 & 12.51 \textpertenthousand & 1.0 & 0.99 & 1.0 & 1.0 & 0.94 & \textcolor{OliveGreen}{0.97} \\
        \hline
        Rbot - Scenario 10 & 559 194 & 9.67 \textpertenthousand & 1.0 & 0.98 & 0.99 & 1.0 & 0.90 & \textcolor{OliveGreen}{0.95} \\
        \hline
        Rbot - Scenario 11 & \textcolor{red}{61 551} & 2.76 \textpertenthousand & 1.0 & 1.0 & 1.0 & \textcolor{red}{0.5} & \textcolor{red}{0.33} & \textcolor{red}{0.4} \\
        \hline
        NSIS.ay - Scenario 12 & \textcolor{orange}{156 790} & 10.20 \textpertenthousand & 1.0 & \textcolor{orange}{0.82} & \textcolor{orange}{0.90} & 0.92 & \textcolor{red}{0.41} & \textcolor{red}{0.56} \\
        \hline
        Virut - Scenario 13 & 1 294 025 & 7.57 \textpertenthousand & 1.0 & 1.0 & 1.0 & 1.0 & 0.96 & \textcolor{OliveGreen}{0.98} \\
        \hline
    \end{tabular}
\end{table}

Table \ref{result_summary_rfc} shows that the Random Forest Classifier succeeds in detecting most of the botnets for 8 scenarios out of 13. The poor scores of the 5 other scenarios seem to be related to the size of the dataset.

\subsubsection{Statistic analysis}
\label{statistic_analysis}
The main problem with the results of Table \ref{result_summary_rfc} is that they are only based on one test dataset. Thus, if the size of the dataset is too small, statistical error may be significant. Therefore, a statistic analysis is made on the scenarios which have small datasets.
\begin{table}[H]
    \centering
    \caption[table]{Statistic analysis of the results over 50 random test datasets}
    \label{statistic_analysis_table}
    \setlength\extrarowheight{5pt}
    \begin{tabular}{|c|c|c|c|c|c|c|}
        \hline \multirow{2}{*}{Botnet} & \multicolumn{3}{c|}{Mean (Test)} & \multicolumn{3}{c|}{Standard deviation (Test)} \\
        \cline{2-7}
         & Precision & Recall & $f_1$ Score & Precision & Recall & $f_1$ Score \\
        \hline
        Rbot - Scenario 4 & 1.0 & \textcolor{orange}{0.60} & \textcolor{orange}{0.75} & 0.0 & 0.08 & 0.06 \\
        \hline
        Virut - Scenario 5 & 1.0 & \textcolor{red}{0.45} & \textcolor{orange}{0.59} & 0.0 & \textcolor{red}{0.21} & \textcolor{red}{0.20} \\
         \hline
        DonBot - Scenario 6 & 1.0 & 0.95 & \textcolor{OliveGreen}{0.97} & 0.0 & \textcolor{OliveGreen}{0.03} & \textcolor{OliveGreen}{0.02} \\
        \hline
        Sogou - Scenario 7 & 1.0 & \textcolor{red}{0.42} & \textcolor{orange}{0.52} & 0.0 & \textcolor{red}{0.30} & \textcolor{red}{0.32} \\
        \hline
        Rbot - Scenario 10 & 0.99 & 0.90 & \textcolor{OliveGreen}{0.94} &  0.01 & \textcolor{OliveGreen}{0.02} & \textcolor{OliveGreen}{0.01} \\
        \hline
        Rbot - Scenario 11 & \textcolor{OliveGreen}{0.95} & \textcolor{orange}{0.50} & \textcolor{orange}{0.63} & \textcolor{orange}{0.12} & \textcolor{orange}{0.18} & \textcolor{orange}{0.16} \\
        \hline
        NSIS.ay - Scenario 12 & 0.93 & \textcolor{red}{0.39} & \textcolor{orange}{0.54} & 0.07 & 0.08 & 0.08 \\
        \hline
    \end{tabular}
\end{table}
Table \ref{statistic_analysis_table} qualifies the result of the previous section because it shows that the score means are not as bad as what can be interpreted from Table \ref{result_summary_rfc}. Indeed, the small sizes of the datasets are responsible for a significant statistical deviation. It must be noted that the preprocessing of feature extraction (see Section \ref{feature_extraction}) is not the reason for a lack of data since the original datasets already have few flows (129 832 for the scenario 5, 114 077 for the scenario 7, 107 251 for the scenario 11).\\

We can then notice that the mean precision for each scenario is close to 95\%, and that the mean recall is around 50\% or more. The Random Forest Classifier seems inefficient only on the scenarios 4 and 12 since the statistical deviations are not very important and the dataset sizes are not very small.

\subsubsection{Generalization}
In order to cope with the lack of data in some scenarios, an idea is to train the classifier on one scenario with a large dataset, and then test it on another scenario with a smaller dataset.
\begin{table}[H]
    \centering
    \caption[table]{Result using Rbot - Scenario 3 as training dataset}
    \label{result_train_using_3}
    \setlength\extrarowheight{5pt}
    \begin{tabular}{|c|c|c|c|}
        \hline \multirow{2}{*}{Botnet} & \multicolumn{3}{c|}{Test} \\
        \cline{2-4}
         & Precision & Recall & $f_1$ Score \\
        \hline
        Rbot - Scenario 3 & 1.0 & 0.96 & \textcolor{OliveGreen}{0.98} \\
        \hline
        Rbot - Scenario 4 & \textcolor{red}{0.0} & \textcolor{red}{0.0} & \textcolor{red}{0.0} \\
        \hline
        Rbot - Scenario 10 & \textcolor{red}{0.0} & \textcolor{red}{0.0} & \textcolor{red}{0.0} \\
        \hline
        Rbot - Scenario 11 & \textcolor{red}{0.0} & \textcolor{red}{0.0} & \textcolor{red}{0.0} \\
        \hline
    \end{tabular}
\end{table}

\begin{table}[H]
    \centering
    \caption[table]{Result using Rbot - Scenario 10 as training dataset}
    \label{result_train_using_10}
    \setlength\extrarowheight{5pt}
    \begin{tabular}{|c|c|c|c|}
        \hline \multirow{2}{*}{Botnet} & \multicolumn{3}{c|}{Test} \\
        \cline{2-4}
         & Precision & Recall & $f_1$ Score \\
        \hline
        Rbot - Scenario 3 & \textcolor{red}{0.0} & \textcolor{red}{0.0} & \textcolor{red}{0.0} \\
        \hline
        Rbot - Scenario 4 & \textcolor{red}{0.6} & \textcolor{red}{0.02} & \textcolor{red}{0.05} \\
        \hline
        Rbot - Scenario 10 & 1.0 & 0.90 & \textcolor{OliveGreen}{0.95} \\
        \hline
        Rbot - Scenario 11 & \textcolor{OliveGreen}{1.0} & \textcolor{orange}{0.41} & \textcolor{orange}{0.58} \\
        \hline
    \end{tabular}
\end{table}

\begin{table}[H]
    \centering
    \caption[table]{Result using Virut - Scenario 13 as training dataset}
    \label{result_train_using_13}
    \setlength\extrarowheight{5pt}
    \begin{tabular}{|c|c|c|c|}
        \hline \multirow{2}{*}{Botnet} & \multicolumn{3}{c|}{Test} \\
        \cline{2-4}
         & Precision & Recall & $f_1$ Score \\
        \hline
        Virut - Scenario 5 & \textcolor{red}{0.0} & \textcolor{red}{0.0} & \textcolor{red}{0.0} \\
        \hline
        Virut - Scenario 13 & 1.0 & 0.96 & \textcolor{OliveGreen}{0.98} \\
        \hline
    \end{tabular}
\end{table}

Tables \ref{result_train_using_3}, \ref{result_train_using_10} and \ref{result_train_using_13} show that training the classifier on another scenario does not work at all. This means it is very difficult or impossible to detect new unseen botnets with our dataset and our method. Consequently, in order to detect a botnet, the classifier must have been trained with a dataset containing this botnet beforehand.\\

Another possible explanation of a bad generalization to other scenarios is that each scenario has a botnet whose purpose is different. In Figure \ref{ctu13}, we can see that the characteristics of the botnet attacks are different so it makes sense that training the classifier with another scenario does not give good result.

\subsubsection{Data augmentation}
\label{data_augmentation}
Another idea to cope with the small sizes of the datasets is to use a data augmentation method to build a larger dataset. Here, we are resampling the training data using the bootstrap technique to make it 10 times (see Table \ref{bootstrap10}) and 30 times bigger (see Table \ref{bootstrap30}).
\begin{table}[H]
    \centering
    \caption[table]{Statistic analysis with data resample (Bootstrap: training data $\times 10$)}
    \label{bootstrap10}
    \setlength\extrarowheight{5pt}
    \begin{tabular}{|c|c|c|c|c|c|c|}
        \hline \multirow{2}{*}{Botnet} & \multicolumn{3}{c|}{Mean (Test)} & \multicolumn{3}{c|}{Standard deviation (Test)} \\
        \cline{2-7}
         & Precision & Recall & $f_1$ Score & Precision & Recall & $f_1$ Score \\
        \hline
        Virut - Scenario 5 & 1.0 & \textcolor{OliveGreen}{0.55} & \textcolor{OliveGreen}{0.69} & 0.0 & 0.20 & 0.17 \\
         \hline
        Sogou - Scenario 7 & 1.0 & \textcolor{OliveGreen}{0.55} & \textcolor{OliveGreen}{0.65} & 0.0 & 0.32 & 0.31 \\
        \hline
        Rbot - Scenario 11 & 0.94 & \textcolor{OliveGreen}{0.63} & \textcolor{OliveGreen}{0.74} & 0.12 & 0.18 & 0.14 \\
        \hline
    \end{tabular}
\end{table}
\begin{table}[H]
    \centering
    \caption[table]{Statistic analysis with data resample (Bootstrap: training data $\times 30$)}
    \label{bootstrap30}
    \setlength\extrarowheight{5pt}
    \begin{tabular}{|c|c|c|c|c|c|c|}
        \hline \multirow{2}{*}{Botnet} & \multicolumn{3}{c|}{Mean (Test)} & \multicolumn{3}{c|}{Standard deviation (Test)} \\
        \cline{2-7}
         & Precision & Recall & $f_1$ Score & Precision & Recall & $f_1$ Score \\
        \hline
        Virut - Scenario 5 & 1.0 & 0.57 & 0.70 & 0.0 & 0.20 & 0.17 \\
         \hline
        Sogou - Scenario 7 & 1.0 & 0.55 & 0.65 & 0.0 & 0.33 & 0.32 \\
        \hline
        Rbot - Scenario 11 & 0.94 & 0.64 & 0.74 & 0.12 & 0.19 & 0.15 \\
        \hline
    \end{tabular}
\end{table}
Tables \ref{bootstrap10} and \ref{bootstrap30} show the statistic analysis of the scenarios 5, 7 and 11 using the bootstrap method as resampling. We can see in Table \ref{bootstrap10} that the recall and $f_1$ score are increased by around 10 points each for all of the scenarios. This means that the size of the dataset is the real problem for detecting botnets. However, Table \ref{bootstrap30} shows that this resampling trick cannot be used as much as we want since the scores reach a saturation point, even if the training dataset increases in size with the bootstrap method.\\

\underline{Remark:} The standard deviations remain unchanged because the bootstrap method is only used with the training dataset and not with the test dataset.

\subsubsection{Other algorithms}
In this section, we compare the results of different algorithms on the scenarios where the Random Forest Classifier performs poorly (ie scenarios 4, 5, 7, 11 and 12).

\paragraph{Logistic Regression}~~\\
A Logistic Regression classification is run to detect botnets in the scenarios 4, 5, 7, 11, and 12. The chosen parameters ($C$ and $\text{Weight}_\text{non-botnet}$) are the same as the ones used for the scenarios 1.
\begin{table}[H]
    \centering
    \caption[table]{Result using Logistic Regression}
    \label{result_logreg}
    \setlength\extrarowheight{5pt}
    \begin{tabular}{|c|c|c|c|c|c|c|}
        \hline \multirow{2}{*}{Botnet} &
        \multicolumn{3}{c|}{Training} &
        \multicolumn{3}{c|}{Test} \\
        \cline{2-7}
         & Precision & Recall & $f_1$ Score & Precision & Recall & $f_1$ Score \\
        \hline
        Rbot - Scenario 4 & \textcolor{red}{0.07} & \textcolor{red}{0.07} & \textcolor{red}{0.07} & \textcolor{red}{0.05} & \textcolor{red}{0.05} & \textcolor{red}{0.05} \\
        \hline
        Virut - Scenario 5 & \textcolor{red}{0.45} & 0.91 & \textcolor{red}{0.60} & \textcolor{red}{0.32} & 0.72 & \textcolor{red}{0.43} \\
        \hline
        Sogou - Scenario 7 & \textcolor{red}{0.15} & \textcolor{red}{0.25} & \textcolor{red}{0.18} & \textcolor{red}{0.02} & \textcolor{red}{0.05} & \textcolor{red}{0.03} \\
        \hline
        Rbot - Scenario 11 & \textcolor{red}{0.27} & \textcolor{red}{0.54} & \textcolor{red}{0.35} & \textcolor{red}{0.20} & \textcolor{red}{0.36} & \textcolor{red}{0.22} \\
        \hline
        NSIS.ay - Scenario 12 & \textcolor{red}{0.09} & \textcolor{red}{0.38} & \textcolor{red}{0.14} & \textcolor{red}{0.08} & \textcolor{red}{0.35} & \textcolor{red}{0.13} \\
        \hline
    \end{tabular}
\end{table}
Table \ref{result_logreg} shows that the Logistic Regression cannot detect botnets in complicated scenarios where data are scarce or not representative of malware behaviours.

\paragraph{Gradient Boosting}~~\\
The same scenarios are tested with the Gradient Boosting method with the same parameters as scenario 1.
\begin{table}[H]
    \centering
    \caption[table]{Result using Gradient Boosting}
    \label{result_gboost}
    \setlength\extrarowheight{5pt}
    \begin{tabular}{|c|c|c|c|c|c|c|}
        \hline \multirow{2}{*}{Botnet} &
        \multicolumn{3}{c|}{Training} &
        \multicolumn{3}{c|}{Test} \\
        \cline{2-7}
         & Precision & Recall & $f_1$ Score & Precision & Recall & $f_1$ Score \\
        \hline
        Rbot - Scenario 4 & 1.0 & 0.59 & 0.75 & 0.98 & 0.42 & 0.58 \\
        \hline
        Virut - Scenario 5 & 1.0 & 1.0 & 1.0 & 0.94 & \textcolor{OliveGreen}{0.65} & \textcolor{OliveGreen}{0.73} \\
        \hline
        Sogou - Scenario 7 & 1.0 & 1.0 & 1.0 & \textcolor{red}{0.68} & 0.51 & 0.55 \\
        \hline
        Rbot - Scenario 11 & 1.0 & 1.0 & 1.0 & 0.86 & 0.56 & 0.66 \\
        \hline
        NSIS.ay - Scenario 12 & 1.0 & \textcolor{red}{0.36} & \textcolor{red}{0.53} & 0.98 & \textcolor{red}{0.22} & \textcolor{red}{0.36} \\
        \hline
    \end{tabular}
\end{table}
Table \ref{result_gboost} shows that the results of Gradient Boosting are closed to the Random Forest ones. The scenario 12 performs very badly whereas scenario 5 has better performance.

\paragraph{Dense Neural Network}~~\\
In Table \ref{result_summary_botnet1}, one can notice that the Neural Network performs very well on the scenario 1. So, we can try to train this model on the scenarios where the Random Forest Classifier is not good at detecting botnets.\\

As shown in Section \ref{statistic_analysis}, a statistical analysis needs to be performed to really demonstrate the performance of the Neural Network (statistics based on 50 random test datasets). Moreover, the sampling technique presented Section \ref{data_augmentation} is replaced by an increase in the number of epochs (20 epochs are chosen).

\begin{table}[H]
    \centering
    \caption[table]{Result using Dense Neural Network}
    \label{result_dense_neural_network}
    \setlength\extrarowheight{5pt}
    \begin{tabular}{|c|c|c|c|c|c|c|}
        \hline \multirow{2}{*}{Botnet} &
        \multicolumn{3}{c|}{Training (Mean)} &
        \multicolumn{3}{c|}{Test (Mean)} \\
        \cline{2-7}
         & Precision & Recall & $f_1$ Score & Precision & Recall & $f_1$ Score \\
        \hline
        Rbot - Scenario 4 & 0.91 & \textcolor{red}{0.29} & \textcolor{red}{0.43} & 0.88 & \textcolor{red}{0.26} & \textcolor{red}{0.39} \\
        \hline
        Virut - Scenario 5 & 0.87 & \textcolor{red}{0.49} & \textcolor{red}{0.61} & 0.79 & \textcolor{red}{0.39} & \textcolor{red}{0.49} \\
        \hline
        Sogou - Scenario 7 & \textcolor{red}{0.08} & \textcolor{red}{0.04} & \textcolor{red}{0.06} & \textcolor{red}{0.0} & \textcolor{red}{0.0} & \textcolor{red}{0.0} \\
        \hline
        Rbot - Scenario 11 & 0.85 & \textcolor{red}{0.31} & \textcolor{red}{0.42} & 0.71 & \textcolor{red}{0.26} & \textcolor{red}{0.35} \\
        \hline
        NSIS.ay - Scenario 12 & \textcolor{red}{0.59} & \textcolor{red}{0.07} & \textcolor{red}{0.13} & \textcolor{red}{0.51} & \textcolor{red}{0.05} & \textcolor{red}{0.09} \\
        \hline
    \end{tabular}
\end{table}
Table \ref{result_dense_neural_network} shows the result of the Neural Network on the difficult scenarios. We can notice that a simple dense architecture cannot properly detect complicated botnet behaviours. Perhaps, a Long Short-Term Memory (LSTM) network used with the unprocessed data could better understand how botnets differ from usual communication in the network.

\newpage
\section{Conclusion}
To conclude, our project aimed at building and comparing models that are able to detect botnets in a real network traffic represented by Netflow datasets.\\

After a strong analysis of the data and a lot of reviews of network security papers, relevant features were extracted. Their influence were then studied through a selection process but no feature was poor enough to be left aside for the training part.\\

Then, different algorithms were tested among which a Logistic Regression, Support Vector Machine, Random Forest, Gradient Boosting and a Dense Neural Network. The Random Forest Classifier was chosen to detect botnets in all the other scenarios, resulting in an detection accuracy of \textbf{more than 95\%} of the botnets for 8 out of 13 scenarios.\\

After that, our group focused on increasing the accuracy for the 5 most difficult scenarios. The use of a bootstrap method to increase the amount of data has resulted in detecting \textbf{more than 55\%} of the scenarios 5, 7 and 11. Only the scores of the scenarios 4 and 12 were difficult to improve ($f_1$ score of $0.75$ for scenario 4 and $0.54$ for scenario 12). This is perhaps due to a bad representation of the botnet behaviours with the extracted features or more complex algorithms need to be used (like recursive deep neural networks).\\

The possible improvement of the work presented here would be to try to modify the extracted features with different widths and strides for the time window and explore more hyperparameters for the difficult scenarios. Another idea would be to try to train and test several scenarios at the same time. Finally, unsupervised learning can be tested to detect the behaviour of botnets without using the labels of the data.

\newpage
\nocite{*}
\printbibliography[heading=bibintoc]
\end{document}